%% file: main.tex
\documentclass[10pt,journal,compsoc]{IEEEtran}
\usepackage{algorithm}
\usepackage{algpseudocode}
\ifCLASSOPTIONcompsoc
  \usepackage[nocompress]{cite}
\else
  \usepackage{cite}
\fi

\ifCLASSINFOpdf
  \usepackage[pdftex]{graphicx}
\else
\fi

\usepackage{amsmath}
\usepackage{array}

\ifCLASSOPTIONcompsoc
 \usepackage[caption=false,font=footnotesize,labelfont=sf,textfont=sf]{subfig}
\else
 \usepackage[caption=false,font=footnotesize]{subfig}
\fi

\usepackage{bbding}
\usepackage{url}
\usepackage{pifont}
\usepackage[numbers,sort&compress]{natbib}
\usepackage{natbib}
\usepackage{amssymb}
\usepackage{booktabs}
\usepackage{algorithm}
\usepackage{algpseudocode}
\usepackage[table]{xcolor}
\definecolor{darkgreen}{rgb}{0.0, 0.5, 0.0}
\usepackage{colortbl}

\usepackage{makecell}

\usepackage[pagebackref=false,breaklinks=true,colorlinks=true,bookmarks=false]{hyperref}
\usepackage[normalem]{ulem}
 \usepackage{multirow} 
\usepackage{ragged2e}
\usepackage{adjustbox}
\begin{document}

\title{OmniHD-Scenes: A Next-Generation Multimodal Dataset for Autonomous Driving}

\author{Lianqing~Zheng*,
        Long~Yang*,
        Qunshu~Lin*,
        Wenjin~Ai,
        Minghao~Liu,
        Shouyi~Lu,
        Jianan~Liu,\\
        Hongze~Ren,
        Jingyue~Mo,
        Xiaokai~Bai,
        Jie~Bai,
        Zhixiong~Ma$^{\dagger}$,
        and~Xichan Zhu

\IEEEcompsocitemizethanks{
    \IEEEcompsocthanksitem * Equal Contribution. $^{\dagger}$ Corresponding Author.
    \IEEEcompsocthanksitem The code is available at: \url{https://github.com/TJRadarLab/OmniHD-Scenes}.
    \IEEEcompsocthanksitem The work of Lianqing Zheng, Long Yang, Wenjin Ai, Hongze Ren, Jingyue Mo, Zhixiong Ma was supported by the National Key R\&D Program of China under grant 2022YFB2503404.
    \IEEEcompsocthanksitem Lianqing Zheng, Long Yang, Wenjin~Ai, Shouyi~Lu,  Hongze~Ren, Jingyue~Mo, Zhixiong~Ma, Xichan Zhu are with the School of Automotive Studies, Tongji University, Shanghai 201804, China. E-mail: \{zhenglianqing, yanglong, 2332971, 2210803, tjrhz, 2433081, mzx1978, zhuxichan\}@tongji.edu.cn.
    \IEEEcompsocthanksitem Qunshu Lin is with the College of Computer Science and Technology, Zhejiang University, Hangzhou 310027, China. E-mail: linskysuka@gmail.com.
    \IEEEcompsocthanksitem Minghao Liu is with the 2077AI Foundation, Singapore. E-mail: dreamforever.liu@gmail.com.
    \IEEEcompsocthanksitem Jianan Liu is with the Momoni AI, Gothenburg, Sweden. E-mail: jianan.liu@momoniai.org
    \IEEEcompsocthanksitem Xiaokai Bai is with the College of Information Science and Electronic Engineering, Zhejiang University, Hangzhou 310027, China. E-mail: shawnnnkb@zju.edu.cn.
    \IEEEcompsocthanksitem Jie Bai is with the School of Information and Electrical Engineering, Hangzhou City University, Hangzhou 310015, China. E-mail: baij@zucc.edu.cn.
    }
}

\markboth{IEEE TRANSACTIONS ON PATTERN ANALYSIS AND MACHINE INTELLIGENCE}%
{Shell \MakeLowercase{\textit{et al.}}: Bare Demo of IEEEtran.cls for Computer Society Journals}

\IEEEtitleabstractindextext{

\begin{abstract}
\justifying
    The rapid advancement of deep learning has intensified the need for comprehensive data for use by autonomous driving algorithms. High-quality datasets are crucial for the development of effective data-driven autonomous driving solutions. Next-generation autonomous driving datasets must be multimodal, incorporating data from advanced sensors that feature extensive data coverage, detailed annotations, and diverse scene representation. To address this need, we present OmniHD-Scenes, a large-scale multimodal dataset that provides comprehensive omnidirectional high-definition data. The OmniHD-Scenes dataset combines data from 128-beam LiDAR, six cameras, and six 4D imaging radar systems to achieve full environmental perception. The dataset comprises 1501 clips, each approximately 30-s long, totaling more than 450K synchronized frames and more than 5.85 million synchronized sensor data points. We also propose a novel 4D annotation pipeline. To date, we have annotated 200 clips with more than 514K precise 3D bounding boxes. These clips also include semantic segmentation annotations for static scene elements. Additionally, we introduce a novel automated pipeline for generation of the dense occupancy ground truth, which effectively leverages information from non-key frames. Alongside the proposed dataset, we establish comprehensive evaluation metrics, baseline models, and benchmarks for 3D detection and semantic occupancy prediction. These benchmarks utilize surround-view cameras and 4D imaging radar to explore cost-effective sensor solutions for autonomous driving applications. Extensive experiments demonstrate the effectiveness of our low-cost sensor configuration and its robustness under adverse conditions. The dataset is available at \url{https://www.2077ai.com/OmniHD-Scenes}.
    
\end{abstract}

}

\maketitle

\IEEEdisplaynontitleabstractindextext
\ifCLASSOPTIONpeerreview
\begin{center} \bfseries EDICS Category: 3-BBND \end{center}
\fi
\IEEEpeerreviewmaketitle

\input{sections/01_intro}

\input{sections/02_related_work}

\input{sections/03_dataset}
\input{sections/05_experiments}

\input{sections/07_conclusion}

\section*{Limitation}
A limitation of the current version of OmniHD-Scenes is that the public road data is collected from a single geographical region (Changsha, China). While we have maximized the diversity of scenarios and road types within this extensive area, algorithms trained exclusively on this data may face challenges generalizing to regions with different road structures, traffic patterns, and driving behaviors. To efficiently address this, we plan to implement a targeted collection scheme guided by prior map knowledge to pinpoint distinct road infrastructures and regional traffic statistics to identify unique traffic cultures, thereby maximizing geographical diversity with optimized costs.

\section*{Acknowledgement}
Thanks to Intelligent Connected Vehicle inspection Center (Hunan) of CAERI Co., Ltd and Shanghai Geometrical Perception and Learning Co., Ltd.
\ifCLASSOPTIONcaptionsoff
  \newpage
\fi


\footnotesize
\bibliographystyle{IEEEtran}
\bibliography{refs}
\vskip -1em

\begin{IEEEbiographynophoto}{Lianqing Zheng} received his B.S. degree in vehicle engineering from Harbin Institute of Technology, Weihai, China, in 2019. He is currently pursuing a Ph.D. in vehicle engineering at the School of Automotive Studies, Tongji University, Shanghai, China. He serves as a peer reviewer for journals and conferences including \textit{IEEE T-ITS}, \textit{T-IV}, \textit{RA-L}, and \textit{ICRA}. He has interned as an algorithm developer at Hongjing Drive, NIO, and Abaka AI. His research interests include data closed-loop systems, multimodal fusion, 4D radar perception, and visual-language models.
\end{IEEEbiographynophoto}
\vspace{-5mm}  
\vfill
\vspace{-5mm}  
\begin{IEEEbiographynophoto}{Long Yang} received the B.S. degree in energy and power engineering from Harbin Institute of Technology, Weihai, China, in 2022. He is currently pursuing the M.S. degree in vehicle engineering at the School of Automotive Studies, Tongji University, Shanghai, China. His current research interests include multi-sensor fusion and deep learning.
\end{IEEEbiographynophoto}
\vspace{-5mm}  
\vfill
\vspace{-5mm} 
\begin{IEEEbiographynophoto}{Qunshu Lin} received his B.S. degree in electrical engineering from Zhejiang University, China, in 2018. He is the founder of Abaka AI, which provides a professional data engineering platform and data services. He is also one of the founders of 2077AI, an international community that spearheads AI evolution via open-source initiatives. 2077AI aims to craft landmark open-source datasets while supporting innovative ideas from talented researchers.
\end{IEEEbiographynophoto}
\vspace{-5mm}  
\vfill
\vspace{-5mm} 
\begin{IEEEbiographynophoto}{Wenjin Ai}  received her B.S. degree in Vehicle Engineering from Tongji University, Shanghai, China, in 2023. She is currently pursuing her M.S. degree in Vehicle Engineering at the School of Automotive Studies, Tongji University. Her research interests focus on autonomous driving perception, including multi-sensor fusion and object detection.
\end{IEEEbiographynophoto}
\vspace{-5mm} 
\vfill
\vspace{-5mm} 
\begin{IEEEbiographynophoto}{Minghao Liu} is an algorithm researcher at 2077 AI. His research areas include MLOps and loosely coupled pose estimation frameworks for multi-sensor systems, autoformalization with large language models, and multi-agent systems.
\end{IEEEbiographynophoto}
\vspace{-5mm}  
\vfill
\vspace{-5mm}  
\begin{IEEEbiographynophoto}{Shouyi Lu} received the B.S. degree from Shandong University of Technology in 2019 and the M.S. degree from Jilin University in 2022, all in vehicle engineering. He is currently pursuing the Ph.D. degree in vehicle engineering at the School of Automotive Studies, Tongji University, Shanghai, China. His research focuses on pose estimation based on deep learning, place recognition, point cloud generation, and multi-sensor fusion.
\end{IEEEbiographynophoto}
\vspace{-4mm}  
\vfill
\vspace{-4mm}  
\begin{IEEEbiographynophoto}{Jianan Liu} received his B.Eng. degree in Electronics and Information Engineering from Huazhong University of Science and Technology, Wuhan, China, in 2007. He received his M.Eng. degree in Telecommunication Engineering from the University of Melbourne, Australia, and his M.Sc. degree in Communication Systems from Lund University, Sweden, in 2009 and 2012, respectively. Jianan has over ten years of experience in software and algorithm design and development. He has held senior R\&D roles in the AI consulting, automotive, and telecommunication industries. His research interests include applying statistical signal processing and deep learning for medical image processing, wireless communications, IoT networks, indoor sensing, and outdoor perception using a variety of sensor modalities like radar, camera, LiDAR, WiFi, etc.
\end{IEEEbiographynophoto}
\vspace{-4mm}  
\vfill
\vspace{-4mm} 
\begin{IEEEbiographynophoto}{Hongze Ren} received the M.S. degree from Hunan University, Changsha, China, in 2021. He is currently pursuing the Ph.D. degree in vehicle engineering with the School of Automotive Studies, Tongji University, Shanghai, China. His current research interests include trajectory prediction and planning system design for autonomous vehicles, deep learning, and autonomous vehicles.
\end{IEEEbiographynophoto}
\vspace{-4mm}  
\vfill
\vspace{-4mm}  
\begin{IEEEbiographynophoto}{Jingyue Mo} received the B.S. degree in vehicle engineering from Tongji University, Shanghai, China, in 2024. She is currently pursuing the M.S. degree in vehicle engineering at the School of Automotive Studies,Tongji University, Shanghai, China. Her current research interests is autonomous driving perception and test evaluation.
\end{IEEEbiographynophoto}
\vspace{-4mm}  
\vfill
\vspace{-4mm}  
\begin{IEEEbiographynophoto}{Xiaokai Bai} received his B.S. degree from Zhejiang University in 2023. He is currently pursuing the Ph.D. degree with the College of Information Science and Electronic Engineering, Zhejiang University, China. His research interests are 3D object detection and autonomous driving.
\end{IEEEbiographynophoto}
\vspace{-4mm}  
\vfill
\vspace{-4mm}  
\begin{IEEEbiographynophoto}{Jie Bai} received the B.S. degree in instrument science and technology from Harbin Institute of Technology, Harbin, China, in 1987. He received the M.S. and Ph.D. degrees in mechanical engineering from Hiroshima University, Japan, in 1992 and 1995, respectively. He is currently a Foreign Fellow of the Russian Academy of Engineering and a Professor with the School of Information and Electrical Engineering, Hangzhou City University, Hangzhou, China. He is the Chair of the SAE International Conference Organizing Committee and the Chair of the Technical Expert Committee. He is also the Secretary General of the Intelligent Transportation Branch of the China Automotive Engineering Institute. His current research interests include intelligent automotive environment perception and decision-making, multi-sensor data fusion, deep learning, and signal processing.
\end{IEEEbiographynophoto}
\vspace{-4mm}  
\vfill
\vspace{-4mm}  
\begin{IEEEbiographynophoto}{Zhixiong Ma} is a Lecturer with the School of Automotive Studies, Tongji University, Shanghai, China. His research interests include automotive active and passive safety technologies, and autonomous vehicle perception systems. He has published a series of academic papers in Chinese and international academic journals and obtained several invention patents. He has presided over several corporate and government projects and won third prize in the Science and Technology Progress of the China Automobile Industry Competition in 2005, and third prize in the Shanghai Science and Technology Progress Competition in 2009. 

\end{IEEEbiographynophoto}
\vspace{-4mm}  
\vfill
\vspace{-4mm}  
\begin{IEEEbiographynophoto}{Xichan Zhu} received the Ph.D. degree from the Department of Automotive Engineering of Tsinghua University, Beijing, China, in 1995. He is a Professor and Doctoral Supervisor with the School of Automotive Studies, Tongji University, and the Chair of the Safety Branch of the Chinese Society of Automotive Engineers. His research interests include active and passive automobile safety. He has published many studies in domestic and foreign academic journals, and has presided over and completed a number of enterprise and government projects. In 1996, Dr. Zhu was awarded first prize in the category of Scientific and Technological Progress by the Ministry of Machinery Industry, and first prize in the category of Automobile Industry Scientific and Technological Progress. In 2001, he won second prize in the category of Automobile Industry Scientific and Technological Progress and, in 2004, he won second prize in the same category. 
\end{IEEEbiographynophoto}

\vspace{-4mm}  

\end{document}

%% file: sections/01_intro.tex
\IEEEraisesectionheading{\section{Introduction}
\label{sec:introduction}}
\begin{figure}
    \centering
    \vspace{-0.5cm}
    \includegraphics[width=\linewidth]{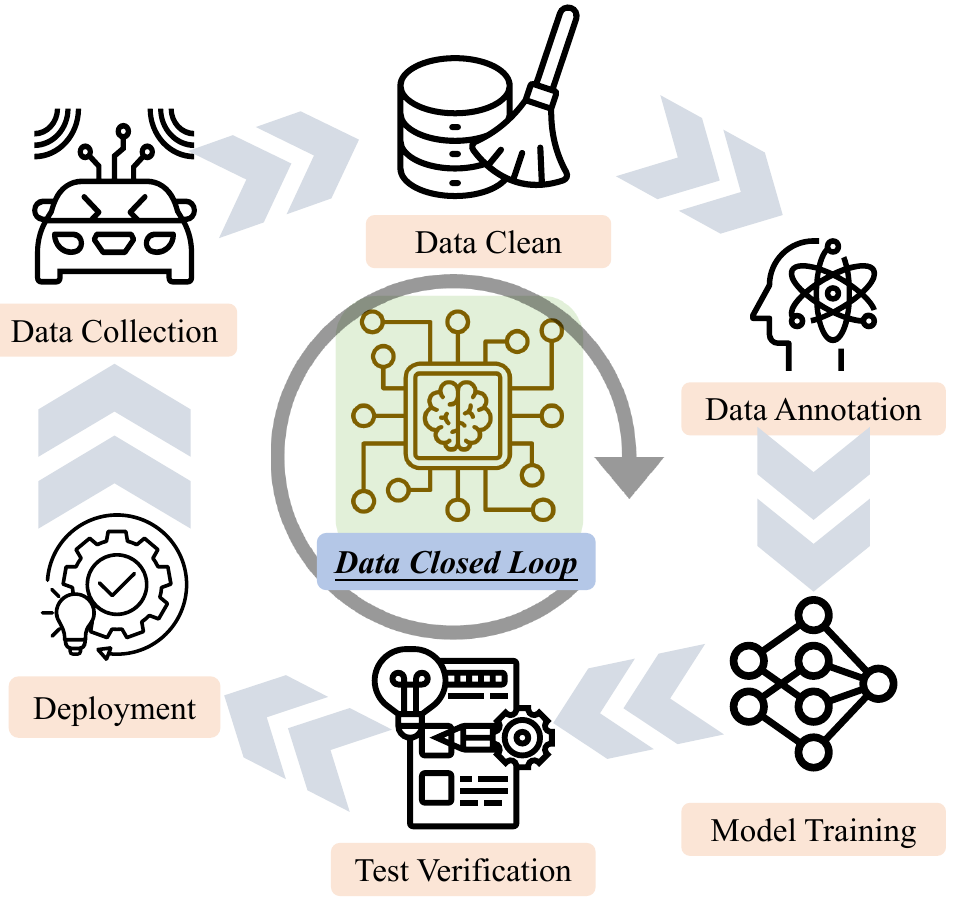}
    \vspace{-0.5cm}
    \caption{Streamlined closed-loop data processing pipeline, primarily comprising data collection, cleaning, annotation, model training, validation, and deployment stages.}
    \vspace{-0.5cm}
    \label{fig:closed-loop}
\end{figure}
\IEEEPARstart{A}{utonomous} driving technology represents the future of transportation and mobility \cite{chen2024smart}, heralding a revolution that will profoundly alter travel methods and economic models. With advancements in computer science, electronic technology, and artificial intelligence, autonomous driving technology is rapidly evolving and permeating various aspects of production and daily life; for example, in unmanned delivery, assisted driving, and robotaxi applications \cite{ZHAOAutonomous_driving_survey}. For data-driven autonomous driving technology, high-quality data and effective algorithms are critical \cite{li2024data}. To address the associated data accumulation challenges, many companies have developed comprehensive data closed-loop platforms. A typical data closed-loop system comprises data collection, cleaning, annotation, model training, validation, and deployment stages, as illustrated in Fig. \ref{fig:closed-loop}. High-quality datasets \cite{liu2024surveydataset} play a crucial role in this pipeline, serving as the foundation for the development and validation of autonomous driving algorithms.

Recently, numerous autonomous driving datasets have emerged. This development has significantly accelerated research advances in related fields \cite{geiger2012KITTI,sun2020WAYMO,caesar2020nuscenes,bansal2020pointillism,mostajabi2020high,bijelic2020seeing,sheeny2021radiate,deziel2021pixset, meyer2019automotive,rebut2022raw,palffy2022multi,zheng2022tj4dradset,paek2022k,zhang2023dual,li2024radset}. For example, the KITTI dataset \cite{geiger2012KITTI}, which was released in 2012, features data from real-world driving scenarios captured using cameras, LiDAR, and GPS/IMU sensors. The Waymo dataset \cite{sun2020WAYMO} incorporates LiDAR and multi-camera sensor data obtained across diverse driving conditions and geographical locations. With its extensive 3D annotations, it has become a benchmark dataset for 3D object detection and tracking tasks. Additionally, nuScenes \cite{caesar2020nuscenes} is an influential large-scale dataset that combines data from LiDAR, six cameras, and five millimeter-wave radar systems. It is widely used for the development and validation of multi-sensor, multi-task surround perception systems, and addresses various perception-system requirements.

However, these datasets have limitations as regards practical application. First, existing datasets do not typically comprise data from high-resolution sensors with comprehensive coverage. Few integrate data from advanced sensors such as high-beam LiDAR, high-resolution multi-view cameras, and multi-view 4D imaging radar systems simultaneously; this shortcoming constrains algorithm development on more sophisticated hardware platforms. Second, as a result of annotation costs and technical constraints, few datasets provide comprehensive annotations including continuous 3D tracking, point-cloud semantic, and occupancy ground-truth labels. Additionally, most datasets lack data for closed test-site scenarios under hazardous conditions; therefore, they provide only limited support for algorithm performance validation and optimization under extreme conditions.

To address these limitations, we present OmniHD-Scenes, a dataset featuring data from omnidirectional sensor coverage captured through six cameras, six 4D imaging radar systems, and one 128-beam LiDAR. The dataset contains data for complex urban scenarios and hazardous test cases captured in closed testing facilities. Using our advanced 4D annotation pipeline, we provide comprehensive 3D bounding box annotations, static point-cloud segmentation labels with full coverage, and automatically generated occupancy ground-truth labels. The contributions of this study can be summarized as follows.

\begin{enumerate}
    \item \textbf{Introduction of a Multi-modal Dataset}. We introduce OmniHD-Scenes, a comprehensive multi-modal dataset, which features emerging 4D imaging radar point clouds for omnidirectional environmental perception. To the best of the authors' knowledge, this is the first open-source dataset with 4D radar point clouds applicable to multiple modern automotive perception tasks (including 3D object detection, multi-object tracking, and occupancy prediction) with multi-view settings. The comprehensive coverage provided by each sensor allows researchers to flexibly explore various sensor combinations and perspectives for multi-modal fusion tasks.
    
    \item \textbf{Superior Data Volume and Scene Coverage}. The dataset surpasses most public datasets as regards both data volume and scene coverage, comprising 1501 continuous 30-s clips that yield more than 450K synchronized frames and 5.85 million synchronized sensor data points. A distinctive feature is the extensive coverage of complex urban traffic scenarios with numerous traffic participants, including challenging conditions such as rainy weather and night scenes. Additionally, we incorporate typical test cases from closed test sites; the acquired data will act as a valuable reference for the development and validation of algorithms such as Autonomous Emergency Braking (AEB).
    
    \item \textbf{Advanced Annotation Pipeline}. We introduce an advanced 4D annotation pipeline that leverages temporal clip data to achieve precise 4D scene reconstruction, thereby enabling semi-automated 3D box annotation and static-scene semantic labeling. Specifically, we combine pre-trained detection models with a refine module to accomplish 3D-object tracking annotation. Therefore, only a few frames are required for completion of keyframe annotations for the entire clip. Through dynamic object removal and the integration of pre-trained segmentation models with clustering algorithms, we achieve precise and efficient static-scene semantic annotation. Additionally, we develop an advanced occupancy generation pipeline that leverages annotation information and non-keyframe data.
    
    \item \textbf{Comprehensive Benchmarks}. We establish comprehensive benchmarks for 3D-object detection and occupancy prediction with corresponding evaluation metrics. We provide baseline models using various sensor modalities, with a primary focus on 4D imaging radar and vision combinations. Hence, the performance of cost-effective sensors applied to these perception tasks is explored.


\end{enumerate}

\input{tables/dataset}

%% file: tables/dataset.tex
\begin{table*}[htbp]
  \begin{center}
  \setlength{\tabcolsep}{3pt}
  \renewcommand\arraystretch{1.1}
  \centering
  
  \caption{Comparison with public datasets for autonomous driving, where ``Scenes'' represents the number of consecutive clips and ``/'' represents unsupported or no statistics. ``Weather'' represents adverse conditions such as rain or fog. ``d/n'' denotes day and night, and ``CTS'' denotes Closed Test Site data. Bolded line represents OmniHD-Scenes.}
  \vspace{-0.5cm}
  \resizebox{\textwidth}{!}{
      \begin{tabular}{cccccccccccccc}
          \toprule[1pt]
            \multirow{2}{*}{{\textbf{Dataset}}} & \multirow{2}{*}{{\textbf{Year}}} & {{\textbf{Anno.}}} & \multirow{2}{*}{{\textbf{Scenes}}}&\multicolumn{3}{c}{{\textbf{Sensors}}}&\multirow{2}{*}{{\textbf{3D Box}}}&\multicolumn{3}{c}{{\textbf{Tasks}}}  & \multirow{2}{*}{{\textbf{Weather}}}  &\multirow{2}{*}{{\textbf{Time}}}&\multirow{2}{*}{{\textbf{CTS}}}\\
            \cline{5-7}\cline{9-11} & &{\textbf{Cover}}&&{\textbf{Camera}}&{\textbf{Radar}}&{\textbf{LiDAR}}  & & {\textbf{3D Det}} & {\textbf{3D Track}} &{\textbf{OCC}}   & \\
          \midrule[0.4pt]
          NuScenes \cite{caesar2020nuscenes} & 2019 &  360° & 1K &6x &5x3d&1x32-beam& 1.4M&$\checkmark$ & $\checkmark$  & $\times$ & $\checkmark$ & d/n & $\times$\\
          Pointillism \cite{bansal2020pointillism} & 2020 & Front  & 48 &1x &2x3d&1x16-beam &/ & $\checkmark$ & $\times$  & $\times$ & $\checkmark$ & d/n & $\times$\\
          Zendar \cite{mostajabi2020high} & 2020 & Front  & 27 &1x &1x3d&1x16-beam &11K&$\checkmark$ & $\times$  & $\times$ & $\times$ & day  & $\times$\\
          Dense \cite{bijelic2020seeing} & 2020 & Front  & / &4x &1x3d&2x64/32-beam &100K& $\checkmark$ & $\times$ & $\times$ & $\checkmark$ & d/n & $\times$\\
          RADIATE \cite{sheeny2021radiate} & 2020 & 360°  & / &2x &1xScan&1x32-beam&n/a(200K-2D)& $\times$ & $\times$ & $\times$ & $\checkmark$ & d/n & $\times$\\      
          PixSet \cite{deziel2021pixset} & 2021 & Front  & 97 &3x &1x3d&2x64-beam&/ & $\checkmark$ & $\checkmark$ & $\times$ & $\checkmark$  & d/n & $\times$\\
          \midrule[0.4pt]
          Astyx \cite{meyer2019automotive} & 2019 & Front  & / &1x &1x4d &1x16-beam &3.1K& $\checkmark$ & $\times$ & $\times$ & $\times$ & day & $\times$\\
          RADIal \cite{rebut2022raw} & 2022 & Front  & 91&1x &1x4d&1x16-beam &n/a(9.55K-2D)& $\times$ & $\times$ & $\times$ & $\times$ & day & $\times$\\
          View-of-Delft \cite{palffy2022multi} & 2022 & Front  & 22 &1x &1x4d&1×64-beam&123K& $\checkmark$ & $\checkmark$ & $\times$ & $\times$ & day & $\times$\\
          TJ4DRadSet \cite{zheng2022tj4dradset} & 2022 & Front  & 44&1x &1x4d&1x32-beam &33.3K& $\checkmark$ & $\checkmark$  & $\times$ & $\times$ & d/n & $\times$\\
          K-Radar \cite{paek2022k} & 2022 & Front  & 58 &4x &1x4d&2×128/64-beam &93.3K& $\checkmark$ & $\checkmark$ &  $\times$ & $\checkmark$ & d/n & $\times$\\
          Dual Radar \cite{zhang2023dual} & 2023 & Front  & 151 &1x &2x4d &1×80-beam &103.2K& $\checkmark$ & $\checkmark$ & $\times$ & $\checkmark$ & d/n & $\times$\\
          L-RadSet \cite{li2024radset} & 2024 & Front  & 280 &4x &1x4d &1×80-beam &133K & $\checkmark$ & $\checkmark$ & $\times$ & $\checkmark$  & d/n& $\times$\\
          MAN TruckScenes \cite{fent2024mantruck} & 2024 & 360°  & 747 &4x &6x4d &6×64-beam &/ & $\checkmark$ & $\checkmark$ & $\times$ & $\checkmark$  & d/n& \Checkmark\\
          \textbf{OmniHD-Scenes(Ours)} & \textbf{2024} & \textbf{360°}  & \textbf{1.5K} &\textbf{6x} &\textbf{6x4d} &\textbf{1×128-beam} &\textbf{514.6K}& \Checkmark & \Checkmark & \Checkmark & \Checkmark & \textbf{d/n}& \Checkmark \\
          \bottomrule[1pt]
      \end{tabular}}
    
    \label{table:overviewdataset}
  \end{center}
\end{table*}

%% file: sections/02_related_work.tex
\section{Related Works}

\label{sec:related_works}
This section reviews relevant literature on autonomous driving datasets, 3D object detection, and semantic occupancy prediction.
\subsection{Autonomous Driving Dataset}

The ongoing development of autonomous driving technology depends on high-quality datasets, which support both the development and validation of algorithms. With recent advancements in sensor technology and data acquisition methods, the importance of research on and application of autonomous driving datasets is increasingly being recognized. Table \ref{table:overviewdataset} lists publicly available datasets that include data from cameras, LiDAR, and millimeter-wave radars. The datasets are categorized by radar type: the top and bottom halves include datasets comprising 3D- \cite{caesar2020nuscenes,bansal2020pointillism,mostajabi2020high,bijelic2020seeing,sheeny2021radiate,deziel2021pixset} and 4D-radar data \cite{meyer2019automotive,rebut2022raw,palffy2022multi,zheng2022tj4dradset,paek2022k,zhang2023dual,li2024radset}, respectively. The NuScenes dataset \cite{caesar2020nuscenes} is widely recognized as a benchmark and comprises data for diverse scenarios with fine-grained annotations, which were obtained via comprehensive sensor configurations. Hence, this dataset provides a foundation for multi-view, multi-sensor fusion for autonomous driving tasks. However, these data were captured via five radar systems with low resolution. Additionally, elevation information is lacking, which constrains the use of this dataset for research on radar perception. The Pointillism \cite{bansal2020pointillism} and Zendar \cite{han2020occuseg} datasets are based on rich sets of sensor types but are small in scale. Dense \cite{bijelic2020seeing} and RADIATE \cite{sheeny2021radiate} contain data on adverse weather and incorporate richer scenarios; however, they lack tracking information. RADIATE contains data from a scanning radar that allowed 360° sensing; however, velocity information is lacking and the data are labeled with a 2D bounding box only, limiting the dataset applicability for 3D perception.

Since the advent of 4D imaging radar, its application to perception research has attracted attention. VoD \cite{palffy2022multi} and TJ4DRadSet \cite{zheng2022tj4dradset} were both presented in 2022. They comprise more annotation information than the alternative datasets Astyx \cite{meyer2019automotive} and RADIAL \cite{rebut2022raw}, and support 3D detection and tracking tasks. However, neither VoD nor TJ4DRadSet includes bad-weather data. K-Radar \cite{paek2022k} provides 4D-radar raw data for bad weather to facilitate research on tasks such as detection and tracking based on the radar spectrum; however, its raw data tensor storage is large, rendering it unsuitable for real-time applications. Recently, the Dual Radar \cite{zhang2023dual} and L-RadSet \cite{li2024radset} datasets were presented, which include more complex urban scene data and incorporate data for rich driving scenarios and inclement weather; these datasets also support 3D detection and tracking tasks. In particular, the Dual Radar dataset is based on data from two different 4D radar sensors to facilitate investigation of their performance differences. MAN TruckScenes \cite{fent2024mantruck}, a dataset released during the same period, includes data from six 4D radar systems and 3D bounding box annotations, and was primarily designed for algorithm development for truck-specific scenarios. In contrast, our OmniHD-Scenes dataset contains data from six 4D radar systems acquired with high-resolution cameras and LiDAR, and also includes data from day/night and bad-weather scenarios to accommodate more complex working conditions. The proposed dataset also includes segmentation and occupancy labels, thereby fully supporting multi-view and multi-sensor tasks.

\subsection{3D Object Detection}

Recent vision-based 3D object detection methods predominantly focus on multi-view camera inputs \cite{philion2020lift, li2022bevformer, huang2021bevdet, wang2022detr3d, liu2023sparsebev}. The introduction of bird’s eye view (BEV) methods has significantly elevated the accuracy of multi-view 3D object detection \cite{li2023delving}. The lift—splat—shot (LSS) method \cite{philion2020lift} is a pioneering depth-based technique that predicts discrete depth probability distributions to lift 2D features into 3D space, generating BEV features through voxel pooling. In contrast, BEVFormer \cite{li2022bevformer} employs a top-down construction approach, utilizing learnable BEV queries to locate spatial features in images and temporal features in previous BEV features. Additionally, several previous studies have utilized sparse query-based architectures to predict object attributes \cite{wang2022detr3d, liu2023sparsebev}.

While vision-based methods show promise, LiDAR provides precise geometric information that addresses camera inherent limitations in 3D detection. Several methods have achieved excellent performance using either LiDAR point clouds \cite{lang2019pointpillars, Centerpoint,shi2020pv} or multimodal fusion approaches \cite{liang2022bevfusion, liu2023bevfusionmit}. Notably, BEVFusion \cite{liang2022bevfusion} effectively integrates LiDAR and camera data through LSS-based camera-to-BEV transformation, enabling robust multimodal detection through flexible sensor fusion.

Additionally, radar sensors offer cost-effective advantages, particularly in adverse weather conditions \cite{yao2023radarsurvey}. While early camera-radar fusion methods focused on ROI-based approaches \cite{nabati2019rrpn, wang2014bionic}, recent techniques have shifted toward BEV representations \cite{kim2023crn, wu2023mvfusion}. With the emergence of 4D radar, studies have explored point cloud-based approaches for 4D radar detection \cite{xu2021rpfa, liu2023smurf}. Additionally, RCFusion \cite{zheng2023rcfusion} leverages RadarPillarNet for hierarchical feature extraction from 4D-radar point clouds, and effectively fuses the BEV features of camera and 4D-radar modalities using interactive attention. LXL \cite{LXL} leverages radar BEV features to infer a 3D occupancy network and assists image-based view transformation using 3D occupancy and estimated depth.

\subsection{Semantic Occupancy Prediction}
Occupancy networks \cite{mescheder2019occupancy} predict occupancy states and attributes for each spatial location, creating a dense world representation that addresses the long-tailed distribution problem of irregular obstacles. Early surround-view camera occupancy networks \cite{li2022bevformer, huang2021bevdet, huang2022bevdet4d} adapted BEV detection approaches with additional channel feature transformation modules for height recovery. In recent approaches, such as FlashOcc \cite{yu2023flashocc} and FastOcc \cite{hou2024fastocc}, the focus is on inference speed optimization. However, BEV-based occupancy prediction inherently introduces misalignment during height recovery. For improved accuracy, 2D--3D spatial attention mechanisms for volumetric feature extraction were incorporated in SurroundOcc \cite{wei2023surroundocc} and Occ3D \cite{tian2024occ3d}; however, this was at the cost of increased computational complexity. 

As more than 70\% of the space is free space, 3D dense representation easily generates resource waste \cite{tang2024sparseocc}. To establish TPVFormer \cite{huang2023tri}, the advantages of explicit and implicit features were combined and BEV was upgraded to orthogonal three-plane tri-perspective view (TPV). The aim was to balance accuracy with speed. For SparseOcc \cite{tang2024sparseocc}, a coarse-to-fine structure was employed, with 3D sparse diffusers being proposed for spatial scene complementation and fine-graining of the essential features.

Characteristically, the accurate spatial ranging capability of LiDAR can be complemented with camera features. In many previous approaches (Openoccupancy \cite{wang2023openoccupancy}, Co-Occ \cite{pan2024co}, and OccGen \cite{wang2024occgen}), the LiDAR feature extraction module was accessed in the vision framework, with geometric-semantic fusion being completed based on interaction of the features of different modalities through a convolutional neural network or attention mechanism. However, there remains a lack of occupancy benchmarks incorporating multi-view 4D radar data.

%% file: sections/03_dataset.tex
\section{OmniHD-Scenes Dataset} 
\label{sec:The_OmniHD-Scenes_dataset}

This section presents a comprehensive overview of our dataset construction pipeline. We provide detailed descriptions of each stage, including the sensor calibration, time synchronization, data collection, data preprocessing, 4D annotation, and occupancy ground-truth generation stages. The dataset statistics are also given.
\subsection{Sensor Suite}

\begin{figure}
    \centering
    \includegraphics[width=\linewidth]{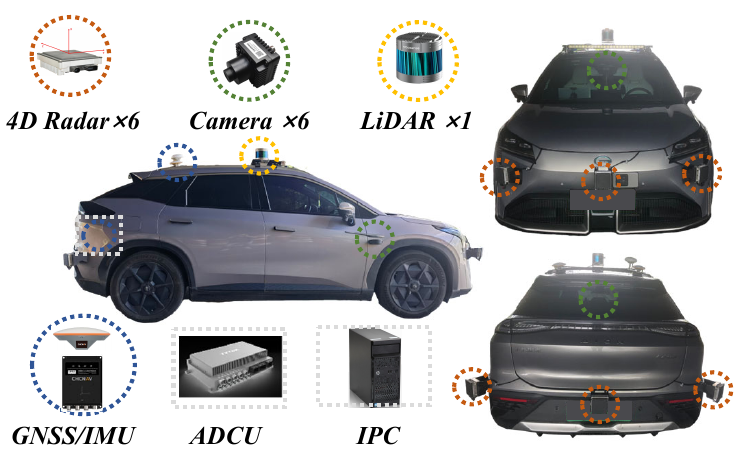}
    \vspace{-0.5cm}
    \caption{Sensor deployment architecture diagram. Multiple views of the vehicle platform are provided to present a comprehensive visualization of the sensor placement.}
    \label{fig:sensor_setup}
\end{figure}

\input{tables/sensor_details}
We employed a variety of high-resolution sensors and meticulously engineered their mounting layout to ensure comprehensive coverage of the vehicle surroundings. The sensor deployment locations and their  specifications are reported in Fig. \ref{fig:sensor_setup} and Table \ref{tab:sensor_setup}, respectively.  These are described as follows.
\begin{itemize}
    \item \textbf{LiDAR}: A high-resolution 128-beam LiDAR, strategically positioned on the vehicle roof, delivers precise three-dimensional point cloud data with exceptional spatial accuracy and an expansive field of view.
	
    \item \textbf{Cameras}: High-resolution 8MP cameras are strategically integrated within the front and rear windshields for high-definition visual data acquisition. Additionally, four 2MP surround-view cameras, mounted on the front fenders, provide comprehensive peripheral visual coverage.
	
    \item \textbf{4D Imaging Radars}: The 4D radar systems are strategically positioned at the front, rear, and four corners of the vehicle. They provide real-time measurements of surrounding objects, including range, azimuth, elevation, and velocity data, thereby generating high-resolution point cloud data.
	
    \item \textbf{Inertial Navigation System}: Positioning and orientation antennas are symmetrically mounted on both sides of the vehicle. These operate in conjunction with an inertial measurement unit (IMU) installed in the trunk to deliver precise pose information for the ego vehicle.
	
\end{itemize}

\begin{figure}
    \centering
    \includegraphics[width=\linewidth]{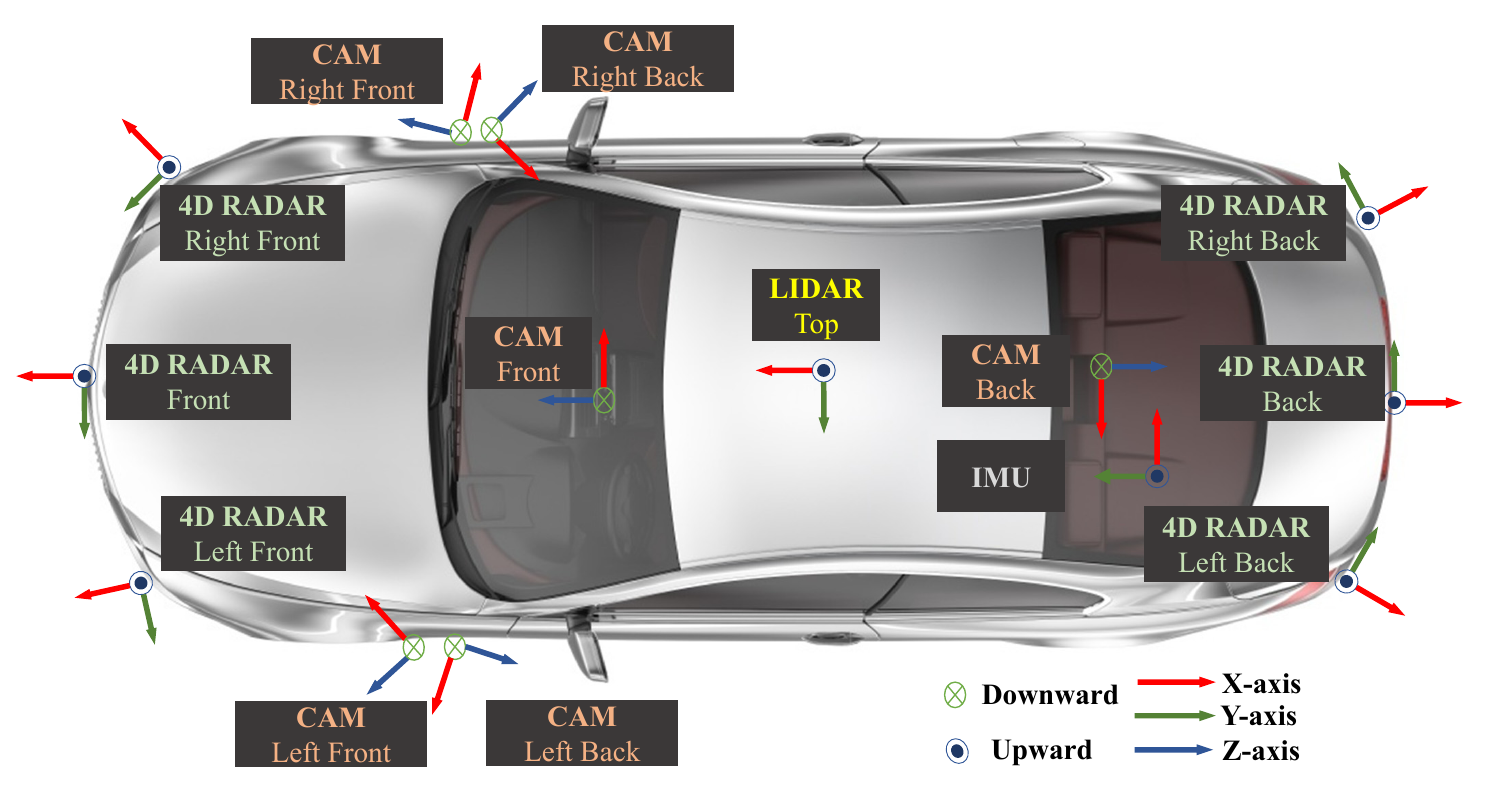}
    \vspace{-0.5cm}
    \caption{Sensor coordinate system. The red, green, and blue arrows  represent the x- y-, and z-axes, respectively. The origin of the vehicle coordinate system is located at the center of the rear axle.}
    \label{fig:sensor-coor}
\end{figure}
\subsection{Sensor Calibration}

\begin{figure}
    \centering

    \includegraphics[width=\linewidth]{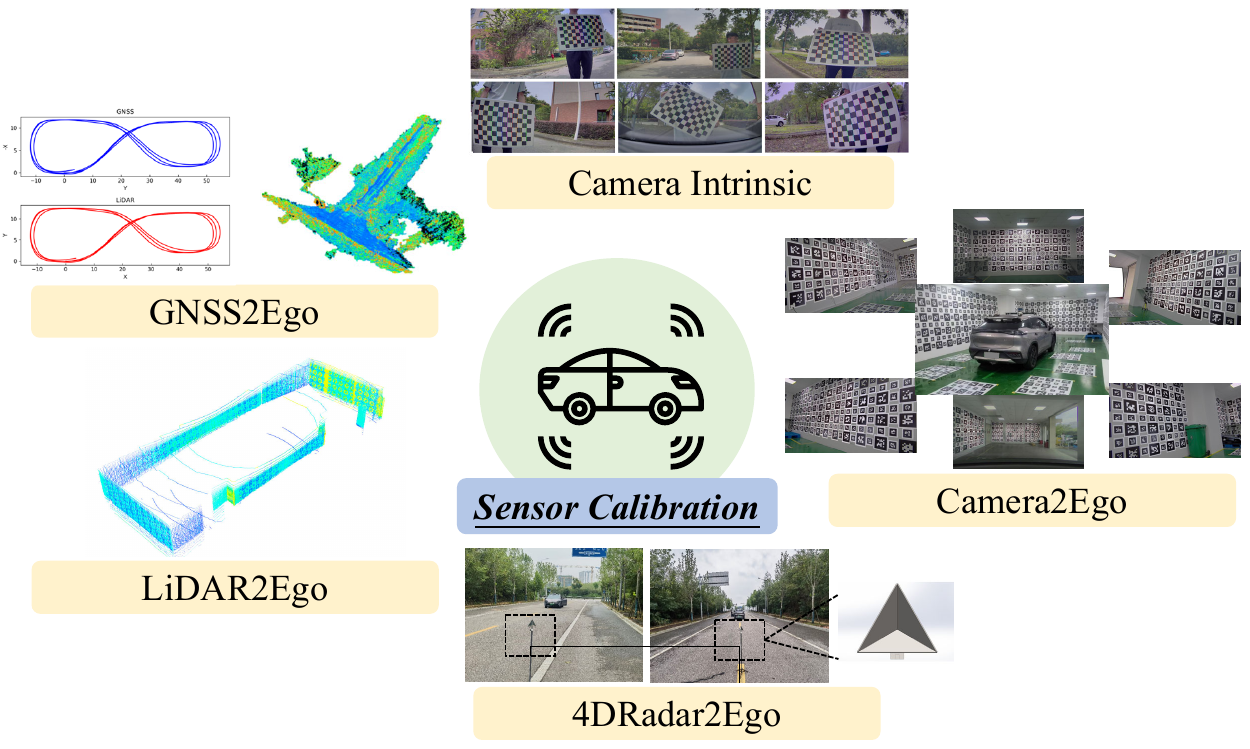}
    \vspace{-0.5cm}
    \caption{Sensor calibration. The process primarily involves calibration of the intrinsic parameters of the cameras and the extrinsic parameters of the cameras, LiDAR, 4D radar, and GNSS.}
    \label{fig:Sensor-Calibration}
\end{figure}

To ensure effective multi-sensor data fusion, we precisely calibrated all sensors. The sensor coordinate system is illustrated in Fig. \ref{fig:sensor-coor}. In this setup, the origin of the ego-vehicle coordinate system is located at the center of the vehicle rear axle. The transformation matrices describe how the other sensor coordinates relate to the origin. Fig. \ref{fig:Sensor-Calibration} shows the calibration diagram. The calibration process was as follows.
\begin{itemize}
    \item \textbf{LiDAR to Ego}: We precisely positioned the vehicle at a designated spot in the calibration room. The extrinsic LiDAR parameters were then obtained by registering the LiDAR point cloud with the high-precision 3D reconstructed scene of the calibration room.
	
    \item \textbf{Camera Intrinsic}: We employed the Zhang calibration method \cite{zhangzhengyou} and used a checkerboard pattern to calibrate the intrinsic parameters and distortion coefficients of the cameras.
	
    \item \textbf{Camera to Ego}: The extrinsic parameters of the cameras were computed using the PnP \cite{pnp} algorithm, by detecting QR code corners in the camera images and utilizing the reconstructed point cloud of the calibration room. To ensure calibration accuracy, we carefully verified the results of LiDAR point cloud projection onto the images.
	
    \item \textbf{4D Radar to Ego}:  We first determined the initial translation parameters of each millimeter-wave radar relative to the vehicle coordinate system. We then performed precise radar calibration by deploying corner reflectors at various positions and orientations to accurately align the LiDAR and radar point clouds in the vehicle coordinate system. To validate the calibration accuracy, we examined the alignment performance in scenarios involving strong metallic reflectors such as guardrails.
	
    \item \textbf{INS to Ego}: We first measured the initial transformation matrix of the inertial navigation system (INS) output position relative to the vehicle coordinate system as the initial values for calibration. The vehicle was then driven along the 8-character trajectory on a flat road to collect LiDAR point cloud data and INS readings. The precise transformation matrix was determined using the OpenCalib tool \cite{yan2022opencalib}. 
	
\end{itemize}

\subsection{Time Synchronization}

\begin{figure}
    \centering
    \includegraphics[width=\linewidth]{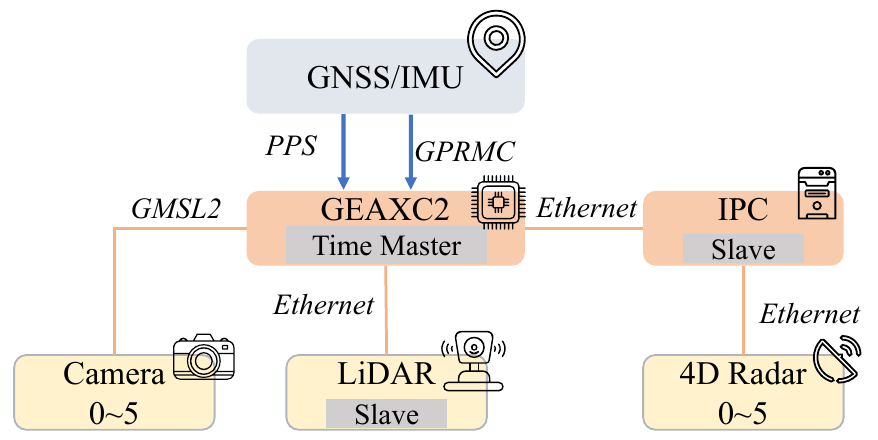}
    \vspace{-0.5cm}
    \caption{Time synchronization system topology. The GEACX2 domain controller acts as a PTP master clock, aligning with the GPS coordinated universal time (UTC). Other devices act as slave nodes to maintain the same time domain.}
    \label{fig:Time-Synchronization}
\end{figure}

To ensure precise timestamp recording of all sensor data within a unified time domain, we implemented a comprehensive time synchronization scheme. The system topology is illustrated in Fig. \ref{fig:Time-Synchronization}.

First, we deployed a precision time protocol (PTP) grandmaster clock on the GEACX2 domain controller, which incorporated an embedded complex programmable logic device (CPLD). The CPLD received pulse per second (PPS) signals and global positioning system recommended minimum data (GPRMC) from an external INS to achieve precise system timing. Hence, synchronization with the GPS coordinated universal time (UTC) was ensured.

We used a Gigabit Multimedia Serial Link 2 (GMSL2) interface to connect the camera system to the controller. Synchronization trigger pulses were generated by the CPLD and sent to the cameras, ensuring synchronized triggering of all six cameras. The LiDAR was connected directly to the domain controller via ethernet, and functioned as a PTP slave clock synchronized to the master clock. The 4D radar systems were connected to a customized industrial computer through ethernet adapters. To ensure accuracy of the 4D-radar data timestamps, the industrial computer also served as a PTP slave clock synchronized with GEACX2.

Through this carefully designed time synchronization scheme, combined with manual verification and correction, we achieved high-precision data synchronization across all sensors in the system.

\subsection{Data Collection}

To maximize the scenario diversity and coverage, we implemented a dual-source data collection strategy through which data were gathered from both public roads and closed test sites. The public road data provided authentic complex scenarios from real-world environments, whereas the closed test sites enabled safe simulation and capturing of data for hazardous situations under controlled conditions.

\subsubsection{Closed Test Site Data}

\begin{figure*}[t]
    \centering
    \includegraphics[width=\linewidth]{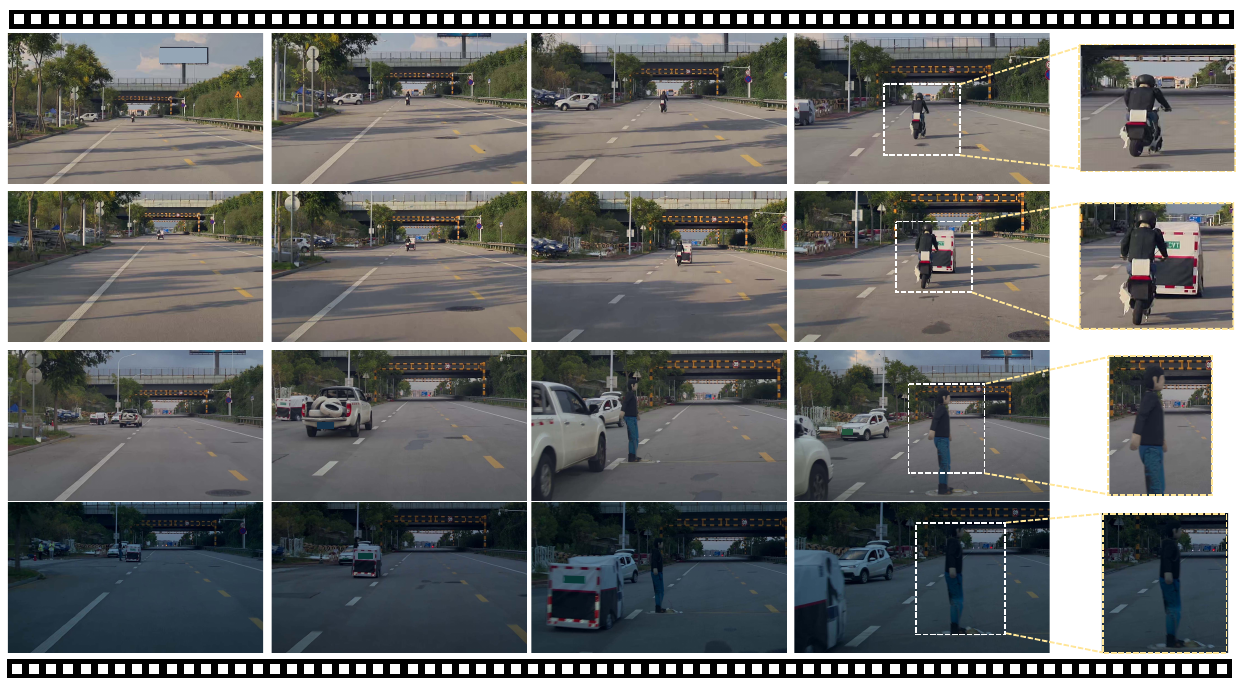}
    \caption{Typical test scenarios. From top to bottom: stationary dummy cyclist; integrated setup combining a dummy cyclist and delivery vehicle; dummy pedestrian crossing under daylight conditions; and dummy pedestrian crossing in nocturnal conditions. Each scenario was systematically designed to incorporate multiple parameter configurations.}
    \label{fig:Closed-Test-Site-Data}
\end{figure*}  

Closed-site testing involves the evaluation of autonomous driving capabilities in controlled, repeatable scenarios created within dedicated testing facilities. These scenarios combine real and simulated traffic participants (such as dummy vehicles and pedestrians) with authentic traffic elements (including roadways, traffic signs, and weather conditions) on closed test sites. In this work, we collected raw sensor data from numerous hazardous test cases to investigate the perception, planning, and control performance of intelligent vehicle strategies under closed test-site conditions. Fig. \ref{fig:Closed-Test-Site-Data} illustrates several typical test cases. The test scenarios included, but were not limited to, the following categories.

\begin{itemize}
    \item \textbf{Longitudinal Braking Scenarios}. These scenarios comprised two primary categories: emergency braking for stationary obstacles (including dummy pedestrians, dummy vehicles, and heterogeneous objects) and emergency braking during car-following situations. All scenarios were implemented at various vehicle speeds, with three time-to-collision (TTC) levels set at 3, 2, and 1.8 s.
	
    \item \textbf{Crossing Scenarios}. These scenarios included crossings by dummy cyclists and pedestrians under various occlusion conditions. All scenarios were implemented at different vehicle speeds, with three TTC thresholds set at 3, 2, and 1.8 s.
	
    \item \textbf{Constant-Speed Scenarios}. These scenarios involved cases in which different object types were placed in adjacent lanes, while the ego vehicle passed at various constant speeds.
	
\end{itemize}

\subsubsection{Public Road Data}
Public-road data collection was primarily conducted in the Changsha Autonomous Driving Demonstration Zone, with the accumulated test mileage exceeding 1,000 km. The data were collected across diverse environmental conditions, spanning different weather patterns, time periods, and complex traffic flows. 

\begin{itemize}
    \item \textbf{Diverse Road Types}. Data for diverse road types, including urban roads, overpasses, and highways, were included, which were collected under both dry and wet conditions. 
	
    \item \textbf{Various Weather Conditions}. Data for driving scenarios under diverse weather conditions, including sunny, cloudy, and rainy days, were collected. This comprehensive collection offers valuable opportunities to assess sensor performance in adverse environments, and is expected to support the development of more robust algorithms.
	
    \item \textbf{Extensive Time Coverage}. Data were obtained for diverse temporal periods, spanning both well-illuminated daytime and challenging low-light nocturnal conditions. This comprehensive coverage will enable researchers to enhance the robustness of perception algorithms across varying illumination scenarios.
	
    \item \textbf{Complex Traffic Flows}. Data were obtained for extensive high-density traffic scenarios involving diverse road participants. These participants exhibited various driving behaviors, including lane changes and emergency braking maneuvers.
	
    \item \textbf{Heterogeneous Driving Behaviors}. Data were collected for a diverse range of driving styles and various typical driving trajectories. Thus, the collected data offer abundant training samples for end-to-end imitation learning.
	
\end{itemize}

\subsection{Data Preprocessing}
The data preprocessing pipeline primarily involved data format conversion, data slicing and synchronization, scene selection, and anonymization.

\subsubsection{Data Format Conversion}
The raw data primarily comprised ROSbag packages (containing LiDAR and 4D radar data), MP4 videos, and TXT files (containing pose data). We converted the data from each sensor into a single-frame format using the original frequency and named each frame with UNIX timestamps for consistency.

The LiDAR data were stored as binary files, which contained the spatial coordinates, intensity, ring number, and timestamp for each point. Each point was transformed to the vehicle coordinate system using extrinsic parameters. Image data were stored in JPG format at native resolution. Similarly, 4D-radar data were stored as binary files, which contained the spatial coordinates, relative radial velocity, amplitude ($Power$), and signal-to-noise ratio ($SNR$). The INS data comprised position and orientation information, velocity components, as well as three-axis measurements from gyroscopes and accelerometers. To enhance the data security, the coordinates were transformed from WGS84 into a local East-North-Up (ENU) coordinate system.

\subsubsection{Data Slicing and Synchronization}
The raw sensor data were precisely sliced into 30-s independent clips based on the timeline. By counting the number of frames from each sensor within each clip and considering the operational frequency of each sensor, we effectively filtered out clips with missing data. These gaps typically resulted from sensor malfunctions or specific manual operations. Following rigorous screening, we ultimately obtained 1,501 complete and valid data clips. 

To synchronize sensors with varying frequencies, we implemented a 10-Hz sampling strategy using LiDAR timestamps as reference points. We selected frames from other sensors closest to these references and matched their ego poses to account for motion between synchronized moments. The synchronized sensor frames and their corresponding poses were systematically preserved within the synchronization file, ensuring precise data alignment and accessibility for further analysis.

\subsubsection{Scene Selection and Anonymization}
We manually annotated key elements of each clip, such as the weather conditions, time of day, and road surface status. Additionally, we classified each sequence as either stationary or dynamic based on the ego-vehicle velocity. To balance the annotation costs and data quality, we carefully selected 200 representative clips for detailed annotation, while ensuring comprehensive coverage of diverse scene types.

For traffic-heavy urban environments, we implemented privacy protection measures by blurring pedestrian faces and vehicle license plates. Using advanced object-detection algorithms \cite{chen2021yolo}, we identified sensitive areas and blurred them.


\subsection{4D Annotation}

\begin{figure}
    \centering
    \includegraphics[width=\linewidth]{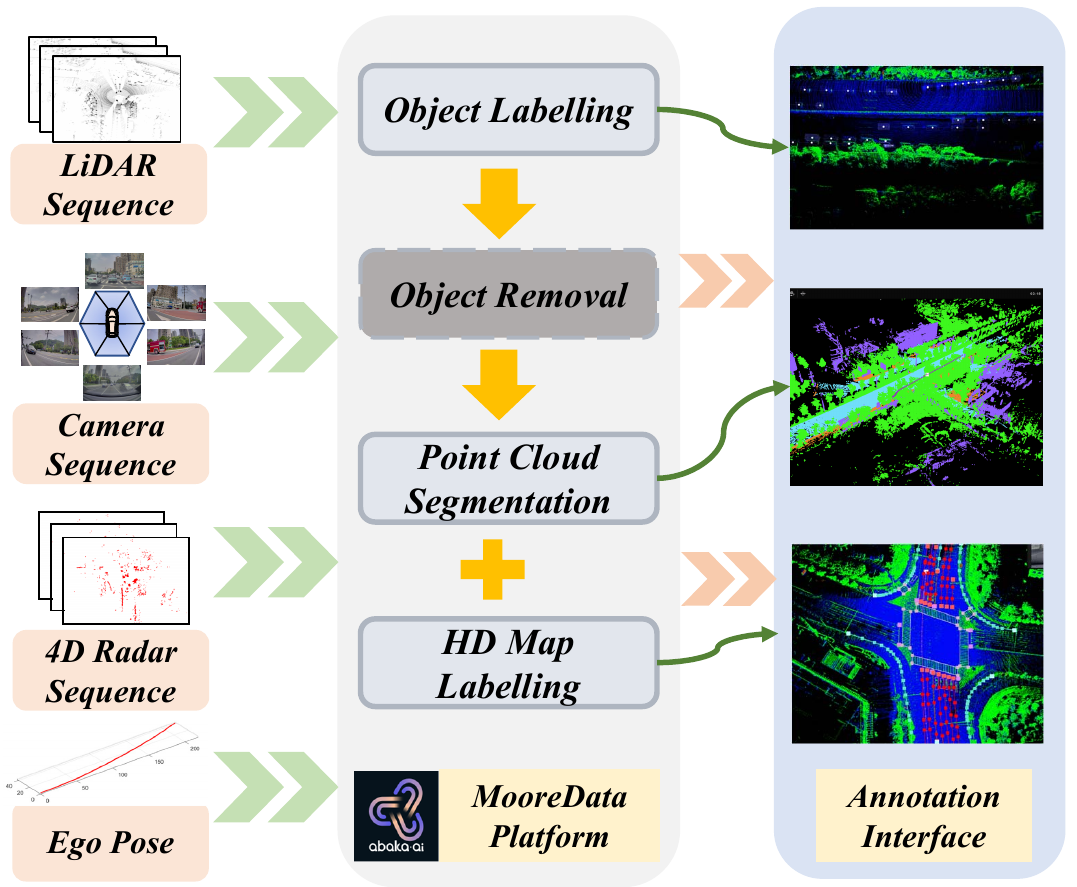}
    \vspace{-0.5cm}
    \caption{4D annotation pipeline and tool interface. The platform processes continuous sensor data and pose information from clips to implement 3D object tracking annotation. Following object removal, the platform reconstructs the static scene, enabling scene-level point-cloud semantic segmentation and map element annotation.}
    \label{fig:4D-annotation}
\end{figure}

Since Tesla introduced the concept of 4D annotation, it has become a pivotal component of the data closed-loop process. This annotation technique utilizes poses to establish temporal relationships and represents traffic participants and road information over a period using dense point cloud reconstruction. Compared to traditional 3D annotation methods, the reconstructed map is denser, exhibits stronger global consistency, offers enhanced visual effects, significantly reduces repetitive tasks, and utilizes more prior information to ensure data reliability. Utilizing 4D tools for data generation can substantially lower data production costs and enhance data quality. In this study, we implemented semi-automatic 4D annotation using the \textit{MooreData} platform solution, with the data being processed in clips. The pipeline and interactive interface for \textit{MooreData} are illustrated in Fig. \ref{fig:4D-annotation}.

First, we applied motion compensation to each LiDAR frame based on the INS data. Subsequently, using pose information from each moment, we performed 4D reconstruction on key frames to obtain a denser point cloud scene within a local coordinate system. Here, the clip start point served as the origin. Next, we annotated 3D tracking boxes for each object; this is referred as ``Object Labelling’’ in Fig. \ref{fig:4D-annotation}. To ensure annotation accuracy, we carefully verified the object trajectories within the clip, thereby preventing problems such as missed labels, ID errors, or size inconsistencies. Subsequently, we removed the annotated objects to obtain a static scene, which was used to annotate the point cloud semantics and static map elements. In the following subsections, we discuss the annotation processes for 3D tracking and point cloud semantics.

\subsubsection{3D Object Annotation}

We presented an efficient semi-automatic 3D object annotation method, the workflow of which was detailed in Algorithm \ref{alg:3d_tracking}. This method began by selecting a uniformly-spaced subset of keyframes from the input LiDAR sequence for manual annotation. For example, in a 60-frame sequence, only 5 keyframes were required. For these selected keyframes, we first generated initial bounding boxes using a pre-trained 3D detection model. These initial boxes were then manually refined and assigned unique identifiers. Subsequently, for the remaining keyframes, we predicted bounding boxes via interpolation, leveraging pose information and the annotated boxes from the selected keyframes. A refinement model, employing an encoder-decoder architecture, iteratively refined these predicted boxes. Each iteration's optimization was evaluated based on the point cloud registration confidence and motion parameters, such as acceleration and angular velocity, of the target object. Optimized bounding boxes were added to the final annotation set only if both the confidence and motion parameters satisfied predefined thresholds; otherwise, the bounding box with the highest confidence was retained. Finally, the algorithm output 3D object annotations for all keyframes. Furthermore, we manually verified the global trajectory of each object and its projections onto the image and point cloud data to ensure high-quality annotations. 

\begin{figure*}[t]
    \centering
    \includegraphics[width=\linewidth]{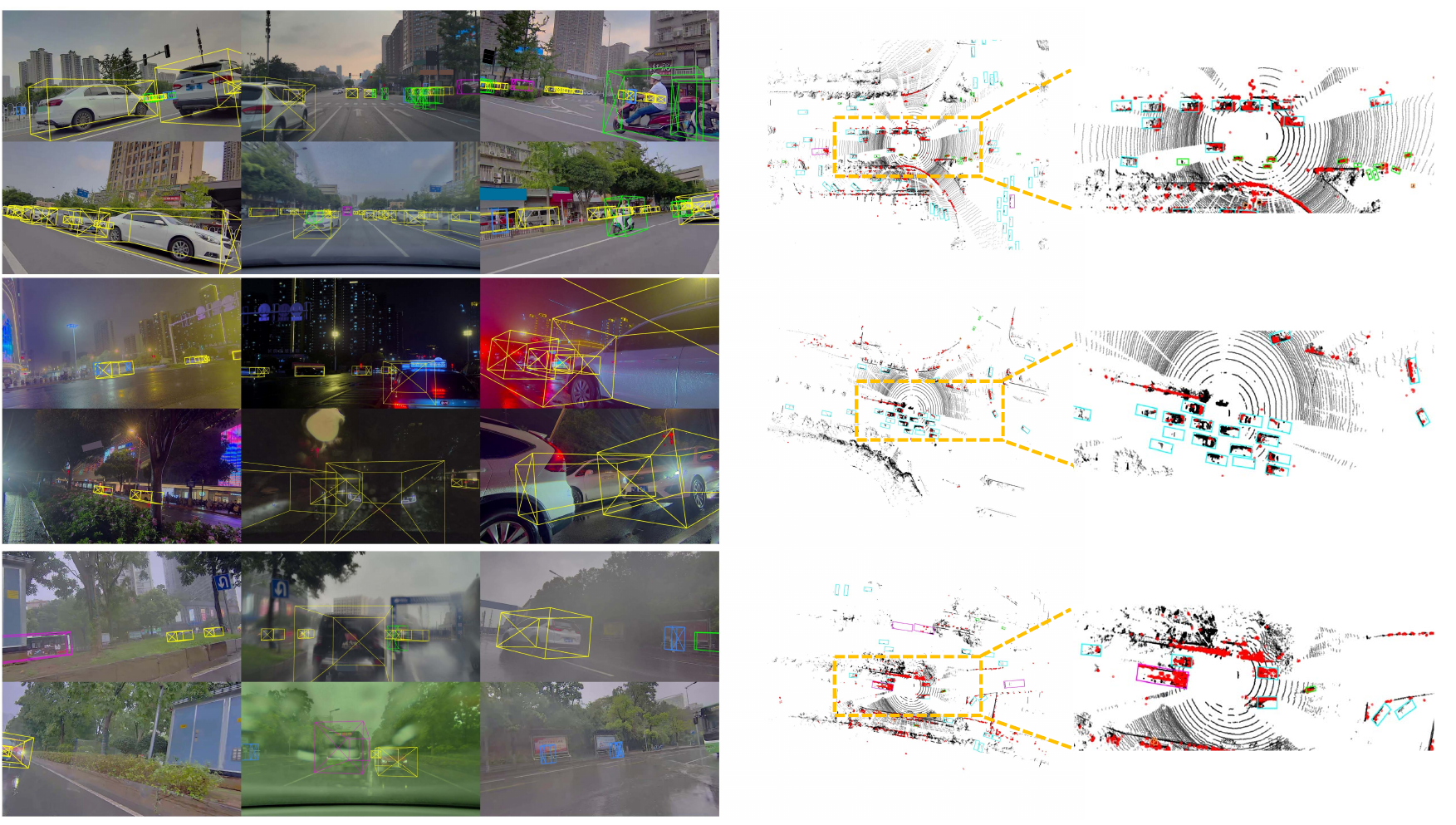}
    \vspace{-0.6cm}
    \caption{Multiple scenes and 3D annotation visualization. (Top to bottom) Bounding boxes projected onto images and 3D space for day, night, and rain scenarios. Black and red points represent LiDAR and 4D-radar data, respectively. }
    \label{fig:3D-Annotation}
    \vspace{0.2cm}
\end{figure*}

Finally, we obtained 3D object annotations for key frames at 2 Hz for each sequence, focusing on objects within 70 m of the front and rear of the ego vehicle, and within 55 m of the left and right. The ground truth annotations are shown in Fig. \ref{fig:3D-Annotation}.

\input{algorithm/3dannotation}

\subsubsection{Static Point Cloud Segmentation Annotation}

\input{algorithm/semantic}
To support point-cloud semantic segmentation and occupancy prediction tasks, we employed a semi-automatic approach to the semantic annotation of static point clouds. The detailed protocol is presented in Algorithm \ref{alg:static_scene_annotation}. First, we obtained a clean static scene by removing all point clouds contained within 3D-object annotation boxes from the reconstructed scene. Next, we leveraged a pre-trained model to segment and extract the road-surface semantic information. The road-surface points were then removed. Subsequently, we employed clustering algorithms to group the remaining points into distinct clusters, thereby enabling efficient annotation through cluster-level classification. Finally, we utilized the lasso tool provided by the MooreData platform for interactive region selection, thereby enabling rapid manual refinement and assignment of semantic labels to point clouds.

To generate high-quality occupancy ground-truth labels, we assigned \textit{"ignore"} labels to noise points in static scenes. These noise points primarily originated from two sources: clustered noise patterns in LiDAR point clouds caused by water splashes from wet road surfaces during rainy conditions, and ghost reflections and other artifacts produced by the inherent limitations of LiDAR sensors.


\begin{figure*}[t]
    \centering
    \includegraphics[width=\linewidth]{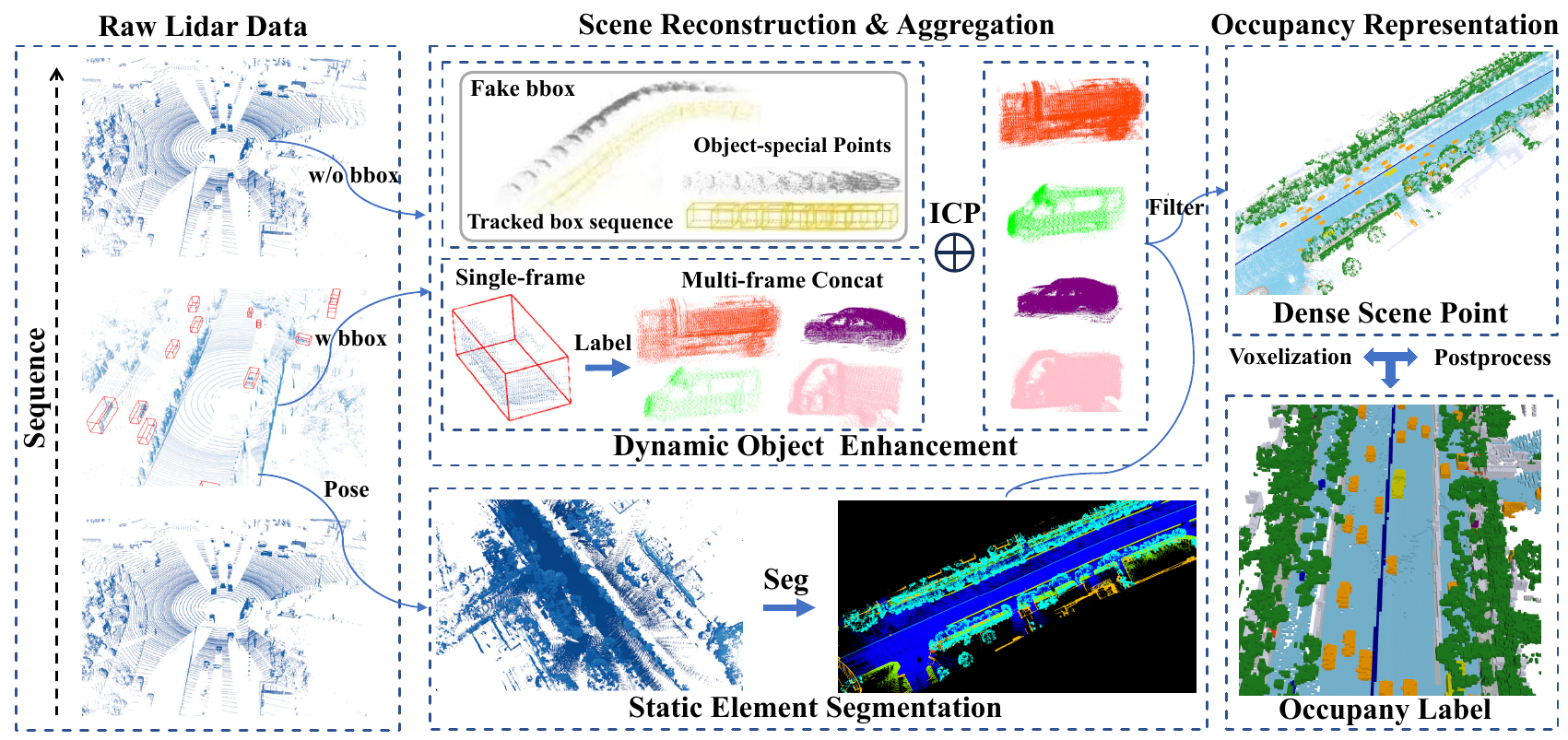}
    \vspace{-0.6cm}
    \caption{Occupancy ground-truth generation pipeline. We first processed keyframe point clouds to separate dynamic objects and static elements using 3D annotations. Non-keyframe objects were extracted through box interpolation. Static elements were transformed to global coordinates and semantically labeled, whereas dynamic objects were enhanced through iterative closest point (ICP) refinement with tracking IDs. Finally, we combined the locally transformed static scene with enhanced objects and applied voxelization to generate occupancy labels, thereby obtaining the reliable ground truth for autonomous driving perception.}
    \label{fig:occgt}
    \vspace{0.2cm}

\end{figure*} 

\begin{figure*}[htbp]
    \centering
    \includegraphics[width=\linewidth]{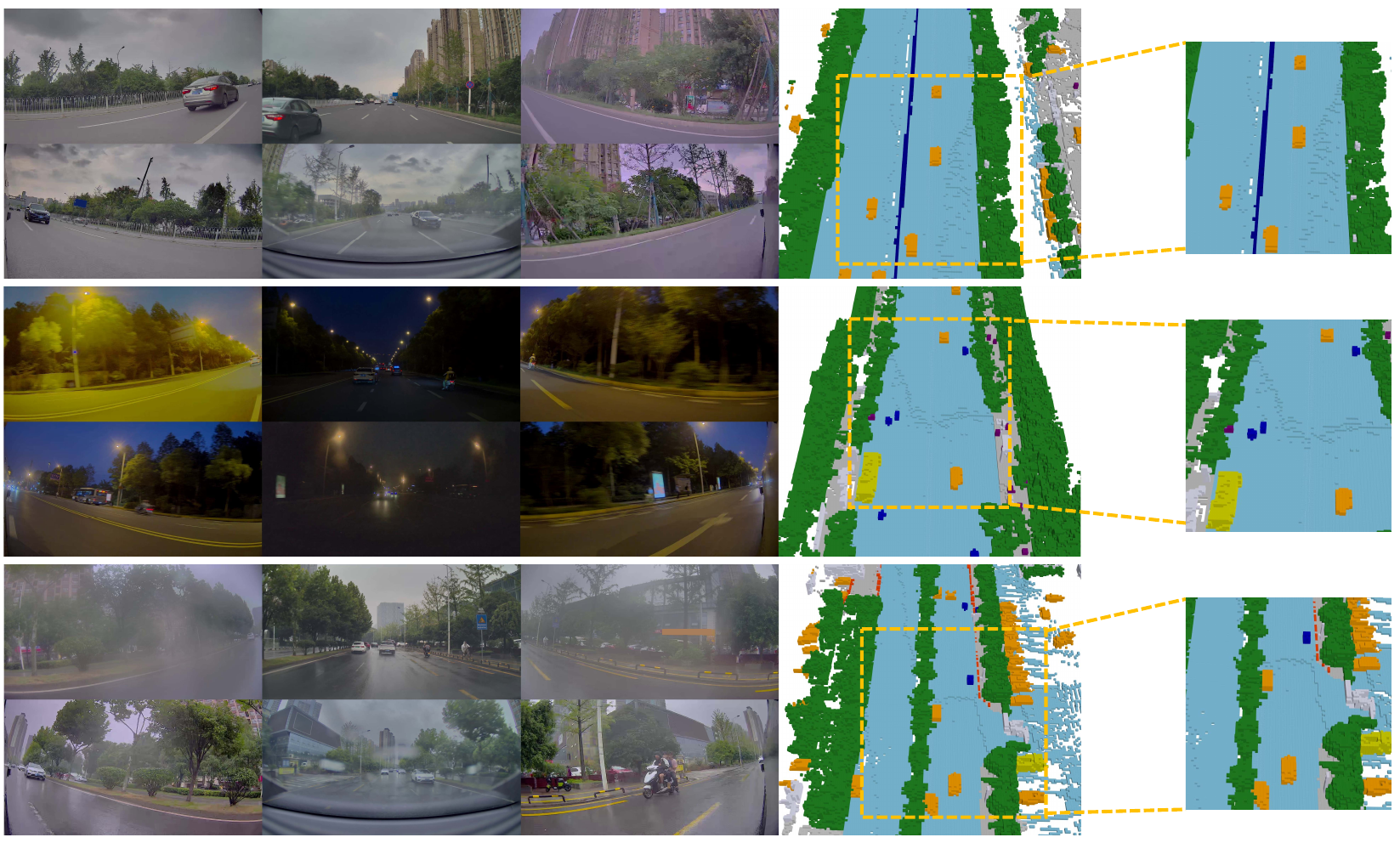}
    \vspace{-0.6cm}
    \caption{Multiple scenes and semantic occupancy visualization. (Top to bottom) Images and occupancy ground truth for daytime, night, and rainy scenes, respectively. The occupancy ground truth contained richer semantic labels and clear object contours. The correspondence between colors and semantic labels can be found in Table \ref{tab:occ_performance}.}
    \label{fig:OCC-Vis}
    \vspace{0.2cm}
\end{figure*} 

\subsection{3D Semantic Occupancy Ground Truth Generation}

\input{algorithm/occ_gt_pipeline}
3D occupancy prediction enables detailed scene understanding by estimating both the state and semantics of each spatial position in the environment. As occupancy benchmarks for surround-view 4D radar were absent, inspired by \cite{wei2023surroundocc}, we developed a high-quality ground-truth generation pipeline for the OmniHD-Scenes dataset, as illustrated in Fig. \ref{fig:occgt}. In contrast to conventional approaches that rely on frame-by-frame panoramic point-cloud segmentation, we leveraged both keyframe 3D annotation and non-keyframe data to enhance the object reconstruction density. The detailed pipeline is outlined in Algorithm \ref{alg:occupancy_gt_pipeline}. 

For key-frame point clouds, we utilized labeled 3D boxes to separate objects from static elements. For unlabeled non-keyframe point clouds, we generated a long-term coarse box sequence based on adjacent keyframes with uniform velocity assumption; this sequence was then used for object extraction. Extracted object point clouds were spatially transformed to object-centric coordinates, and point clouds with sequentially identical tracking IDs were aggregated. Non-keyframe dynamic objects were also used for data augmentation. However, for simple coordinate transformations, spatial alignment errors accumulated. Therefore, ICP matching was employed to eliminate interpolated frame errors during stitching. All keyframe static point clouds were transformed to the global coordinate system to generate the reconstructed static scene, which was then semantically labeled with annotation information. The static scene was transformed back to the local coordinates of each keyframe using inverse pose transformation, and combined with enhanced objects of the current frame to obtain dense scene representations. Radius filtering was then applied to reduce the noise that may have been present during the multi-frame accumulation process.

In previous methods, point-cloud densification was attempted using techniques such as Poisson reconstruction \cite{kazhdan2006poisson} and VBDFusion \cite{vizzo2022vdbfusion}. However, these approaches are primarily suited to sparse LiDAR data and may introduce reconstruction artifacts such as uneven road surfaces. Therefore, we used voxelization and the nearest neighbor algorithms only to assign the occupancy states and semantic labels. The occupancy ground-truth visualization is shown in Fig. \ref{fig:OCC-Vis}.

\subsection{Dataset Statistics}
\begin{figure*}[t]
    \centering
    \includegraphics[width=\linewidth]{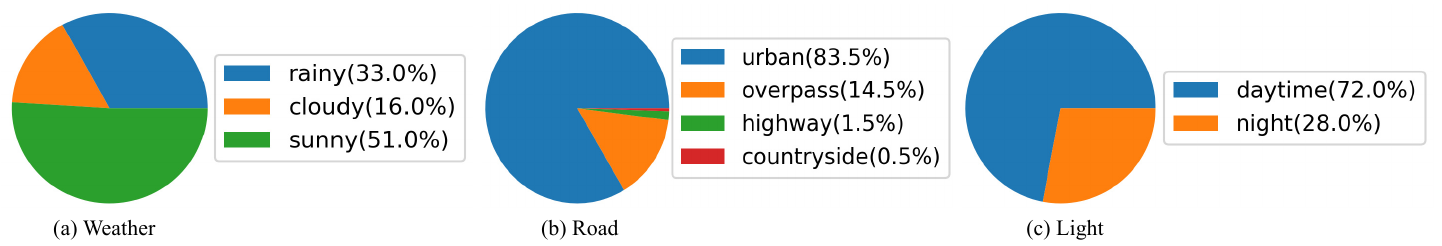}
    \vspace{-0.6cm}
    \caption{Scene element distribution in annotated clips. (a), (b), and (c) show the weather, road-type, and lighting condition distributions. The OmniHD-Scenes dataset emphasizes urban and overpass roads with complex traffic patterns. It is characterized by data pertaining to a significant number of challenging weather conditions, especially rain, and a substantial number of night scenes.}
    \label{fig:Distribution of scene}
\end{figure*} 

\begin{figure}
    \includegraphics[width=\linewidth]{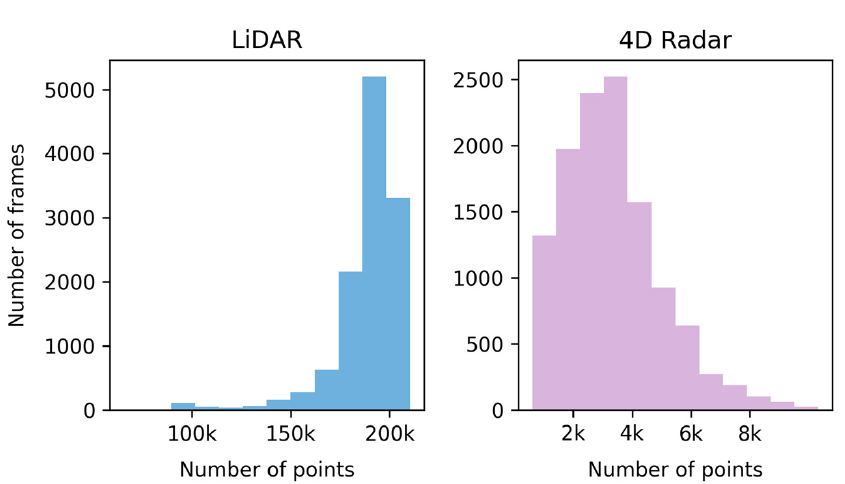}
    \vspace{-0.6cm}
    \caption{Keyframe point-cloud counts. Most LiDAR point-cloud counts are between 180K and 210K, whereas most 4D-radar point-cloud counts are between 2K and 4K.}
    \label{fig:number_points}
\end{figure}
\begin{figure}
    \centering
    \includegraphics[width=\linewidth]{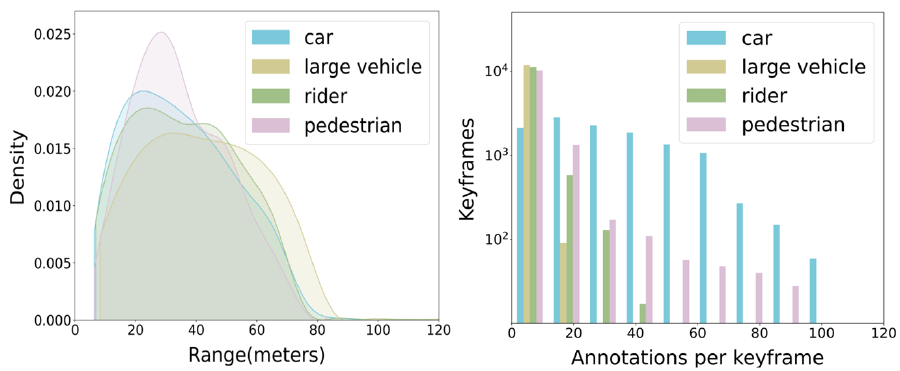}
    \vspace{-0.6cm}
    \caption{Distance distribution and object count statistics. (left) Object distances relative to ego vehicle. (right) Per-frame counts of different categories (cars, large vehicles, riders, and pedestrians). The high density of car annotations (often exceeding 60 per frame) highlights the scene complexity.}
    \label{fig:objects_per_frame}
\end{figure}

The OmniHD-Scenes dataset contains 1501 clips, each lasting approximately 30s, comprising 450k+ synchronized frames and 5.85M+ synchronized sensor data. To date, we have annotated a total of 200 clips, 11921 key frames, and 514619 3D objects, including car, large vehicle, rider, and pedestrian. 

We analyzed the scene-element distribution within the annotated sequences to highlight the dataset diversity, as depicted in Fig. \ref{fig:Distribution of scene}. The OmniHD-Scenes dataset predominantly includes data for urban areas and overpass roads with complex traffic patterns, and spans a range of weather conditions and different times of day. Significantly, it comprises data for a substantial number of challenging scenarios, such as rainy conditions (33\%) and night scenes (28\%), offering valuable opportunities for the development and evaluation of more robust perception algorithms.

Moreover, a notable feature of OmniHD-Scenes is its incorporation of  multi-view 4D radar. Fig. \ref{fig:number_points} illustrates the point-cloud quantity distributions per frame for both the LiDAR and 4D-radar data. The LiDAR point-cloud counts per frame are concentrated between 180K and 210K points, whereas those for the 4D radar primarily range from 2K to 4K points. Therefore, in terms of point-cloud density, the 4D-radar data are conspicuously sparse. This sparseness will inspire researchers to explore more effective ways of leveraging the unique characteristics of 4D radar or to integrate it with other modalities for enhanced perception efficiency.

Fig. \ref{fig:objects_per_frame} presents the 3D annotation distribution. The left figure highlights the distance distribution between the annotated objects and ego vehicle. Pedestrians are typically concentrated in areas closer to the ego vehicle, particularly at intersections or crosswalks, whereas other objects are more widely dispersed within the annotation range. The right figure shows the category count for each keyframe. For the car class, many frames contain more than 60 cars, underscoring the traffic complexity in urban scenes. Moreover, other object categories are also well-represented. This extensive annotation not only enhances the dataset value as regards the training of complex models but also poses challenges that will further the development of object detection techniques, etc.



%% file: tables/sensor_details.tex
\begin{table*}[t]
    \centering
    \caption{Sensor and controller specifications of OmniHD-Scenes. We describe the detailed configuration of the used devices.}
    \vspace{-0.2cm}
    \begin{tabular}{l|l|lp{9.3cm}}
    \toprule
    \textbf{Sensor} & \textbf{Sensor Model} & \textbf{Details} \\ \toprule
    LiDAR & Robosense RS-Ruby-128 ($\times$1) & 128 beams, $360^\circ$ HFOV, $-25^\circ \sim +15^\circ$ VFOV, 10 FPS\\ 
    \midrule
    \multirow{3}{*}{Camera} & SG8-AR0820C ($\times$1) & Front-view, $60^\circ$ HFOV, $33^\circ$ VFOV, 3840$\times$2160 resolution, 30 FPS\\ 
    & SG8-AR0820C ($\times$1) & Rear-view, $120^\circ$ HFOV, $55^\circ$ VFOV, 3840$\times$2160 resolution, 30 FPS\\
    & SG2-AR0233C ($\times$4) & Surround-view, $100^\circ$ HFOV, $53^\circ$ VFOV, 1920$\times$1080 resolution, 30 FPS\\ 
    \midrule
    \multirow{2}{*}{4D Radar} & GPAL Ares-R7861 ($\times$2) & Front/Rear-view, $76 \sim 79$ GHz, $140^\circ$/$20^\circ$ HFOV, $24^\circ$/$10^\circ$ VFOV, 13 FPS\\
    & GPAL Ares-R7861 ($\times$4) & Surround-view, $76 \sim 79$ GHz, $150^\circ$ HFOV, $30^\circ$ VFOV, 13 FPS\\ 
    \midrule
    INS & CGI-430 ($\times$1) &Position accuracy $0.01m(H)/0.02m(V)$, 100 FPS\\ 
    \midrule
    Domain Controller & TZTEK-GEACX2 ($\times$1) & 2$\times$ Jetson AGX Orin Processors, 400TOPS\\ 
    \midrule
    IPC & Customized hp ML10 ($\times$1) & Intel i7-11700k CPU, GeForce RTX 3060 GPU\\ 
    \bottomrule
    \end{tabular}
    \label{tab:sensor_setup}
\end{table*}

%% file: algorithm/3dannotation.tex
\begin{algorithm}[t]
\caption{Semi-Automatic 3D Object Annotation}
\label{alg:3d_tracking}
\textbf{Input:} LiDAR keyframe sequence $\mathcal{S}$, Pre-trained 3D detection model $\mathcal{L}_{pre}$, 
Refinement model $\mathcal{L}_{refine}$, Confidence threshold $\theta_{conf}$, Pose information $\mathcal{P}$

\textbf{Output:} 3D object annotation  $\mathcal{A}_{final}$

\begin{algorithmic}[1]

    \State \textcolor{magenta}{// Keyframe selection and manual refinement}
    \State $\mathcal{K} \leftarrow \text{SelectKeyFrames}(\mathcal{S}) $ 
    \State $\mathcal{B}_{init} \leftarrow \mathcal{L}_{pre}(\mathcal{K}) $ 
    \State $\mathcal{B}_{gt} \leftarrow \text{ManualRefineBBoxwithID}(\mathcal{K},\mathcal{B}_{init})$
    
    \State $\mathcal{A}_{final} \leftarrow \emptyset$ 

    \State \textcolor{magenta}{// Process each frame in the sequence}
    \For{each frame $f_k \in \mathcal{S}$}  
        \State $\mathcal{B}_{tmp} \leftarrow \text{Interpolate}(\mathcal{B}_{gt}, \mathcal{P}, f_k)$
        
        \State $attempt \leftarrow 0$
        \State $max\_conf \leftarrow 0$
        \State $best\_bbox \leftarrow null$
        
        \State \textcolor{magenta}{// Iterative refinement of the bounding box}
        \While{$attempt < 5$} 
            \State $\mathcal{B}_{refine} \leftarrow \mathcal{L}_{refine}(\mathcal{B}_{tmp}, f_k)$ 
            
            \State $conf \leftarrow \text{ICPConfidence}(f_k, \mathcal{K}, \mathcal{B}_{refine}, \mathcal{B}_{gt})$
            \If{$conf > max\_conf$} 
                \State $max\_conf \leftarrow conf$
                \State $best\_bbox \leftarrow \mathcal{B}_{refine}$
            \EndIf
            
            \If{$conf \geq \theta_{conf}$} 
                \State $a_{cal}, w_{cal} \leftarrow \text{CalMotion}( \mathcal{B}_{refine}, \mathcal{A}_{final}, \mathcal{P})$ 
                \If{$a_{cal} \leq a_{thresh}$ \textbf{and} $w_{cal} \leq w_{thresh}$} 
                    \State $\mathcal{A}_{final} \leftarrow \mathcal{A}_{final} \cup \mathcal{B}_{refine}$ 
                    \State \textbf{break} 
                \EndIf
            \EndIf
            
            \State $attempt \leftarrow attempt + 1$
            \If{$attempt = 5$} 
                \State $\mathcal{A}_{final} \leftarrow \mathcal{A}_{final} \cup best\_bbox$
            \EndIf
        \EndWhile
    \EndFor
    \State \textcolor{magenta}{//Post-processing and check}
    \State $\mathcal{A}_{final} \leftarrow \text{Check}(\mathcal{A}_{final})$  
    \State \Return $\mathcal{A}_{final}$
\end{algorithmic}
\end{algorithm}

%% file: algorithm/semantic.tex



       
       
       
       

\begin{algorithm}[t]
\caption{Semi-Automatic Scene Semantic Annotation}
\label{alg:static_scene_annotation}
\textbf{Input:} Reconstructed scene $P$, 3D boxes $B$, Pre-trained segmentation model $M$

\textbf{Output:} Labeled point cloud $P_s$

\begin{algorithmic}[1]
    \State \textcolor{magenta}{// Remove objects' points}
    \State $P_{\text{st}} \leftarrow \text{RemoveObjects}(P, B)$

    \State \textcolor{magenta}{// Road surface segmentation and removal}
    \State $P_r \leftarrow \text{RoadSeg}(P_{\text{st}}, M)$
    \State $P_n \leftarrow P_{\text{st}} \setminus P_r$

    \State \textcolor{magenta}{// Cluster remaining points}
    \State $C \leftarrow \text{Cluster}(P_n)$
    
    \State \textcolor{magenta}{// Assign semantic labels to clusters}
    \For{each $c_i \in C$}
        \State $l_i \leftarrow \text{LabelCluster}(c_i)$
    \EndFor

    \State \textcolor{magenta}{// Manual refinement of labels}
    \State $P_s \leftarrow \text{Refine}(C, l)$

    \State \Return $P_s$
\end{algorithmic}
\end{algorithm}

%% file: algorithm/occ_gt_pipeline.tex
\begin{algorithm}
\caption{Occupancy Ground Truth Generation Pipeline}
\label{alg:occupancy_gt_pipeline}

\textbf{Input:} LiDAR sequences $\mathcal{L}$, Keyframe boxes $\mathcal{B}$, Poses $\mathcal{P}$

\textbf{Output:} Dense occupancy labels $\mathcal{O}$ with semantics

\begin{algorithmic}[1]
    \State \textcolor{magenta}{// Process keyframes and non-keyframes}
    \For{each point cloud $P_t \in \mathcal{L}$}
        \If{$P_t$ is keyframe}
            \State $D_{\text{obj}} \leftarrow \text{ExtractObject}(P_t, \mathcal{B})$ 
            \State $S_{\text{static}} \leftarrow \text{Separate}(P_t, D_{\text{obj}})$
        \Else
            \State $\mathcal{B}^* \leftarrow \text{InterpolateBoxes}(\mathcal{B}, \text{uniform velocity})$
            \State $D_{\text{obj}}^* \leftarrow \text{ExtractObject}(P_t, \mathcal{B}^*)$
        \EndIf 
    \EndFor

    \State \textcolor{magenta}{// Process static scene}
    \State $S_{\text{global}} \leftarrow \text{TransformToGlobal}(S_{\text{static}}, \mathcal{P})$
    \State $S_{\text{labeled}} \leftarrow \text{UseSegLabel}(S_{\text{global}})$
    
    \State \textcolor{magenta}{// Process keyframe objects}
    \For{each $d \in D_{\text{obj}}$}
        \State $D_{\text{agg}} \leftarrow \text{AggregateByID}(d)$
    \EndFor

    \State \textcolor{magenta}{// Align interpolated objects}
    \For{each $d^* \in D_{\text{obj}}^*$}
        \State $d^*_{\text{refined}} \leftarrow \text{ICPAlignment}(d^*, D_{\text{agg}}, ID)$
        \State $D_{\text{agg}} \leftarrow \text{Aggregate}(d^*_{\text{refined}}, d_{\text{agg}})$
    \EndFor

    \State \textcolor{magenta}{// Transform and combine scenes}
    \State $S_{\text{local}} \leftarrow \text{TransformToLocal}(S_{\text{labeled}}, \mathcal{P}^{-1})$
    \State $S_{\text{combined}} \leftarrow \text{CombineScenes}(D_{\text{agg}}, S_{\text{local}})$
    \State $S_{\text{filtered}} \leftarrow \text{RadiusFilter}(S_{\text{combined}})$
    
    \State \textcolor{magenta}{// Generate final occupancy}
    \State $\mathcal{O} \leftarrow \text{NN}(\text{Voxelization}(S_{\text{filtered}}))$
    
    \State \Return $\mathcal{O}$
\end{algorithmic}
\end{algorithm}

%% file: sections/05_experiments.tex
\section{Benchmark Experiments}
\label{sec:exp}


We established robust benchmarks for 3D object detection and 3D occupancy prediction tasks using the OmniHD-Scenes dataset. To facilitate standard benchmarking, the 200 annotated clips are partitioned into training, validation, and test sets at a 6:1:3 ratio. Considering the evaluation inaccuracies caused by scene overlap in online mapping \cite{lilja2024localization}, we ensured that 93.33\% of our splits have no geographic overlap. For the downstream occupancy prediction task, the negligible remaining overlap does not negatively impact the evaluation, as the distinct dynamic objects and varying environmental conditions (e.g., weather and lighting) prevent memorization, thereby establishing a rigorous standard Our aim was to establish strong baseline performance standards that would facilitate and advance future research efforts.

\subsection{3D Object Detection}
\label{sec:3D OD}

\subsubsection{Evaluation Metrics}
For the 3D detection task, four categories must be detected, where each category includes a 3D box (containing the center point, length, width, height, and yaw) and velocity ($v_x$ and $v_y$). We evaluated a detection area extending ±60 m longitudinally and ±40 m laterally relative to the ego vehicle. We followed the nuScenes \cite{caesar2020nuscenes} approach to obtain the mean average precision (mAP) and four mean true positive metrics (mTP): the mean average translation error (mATE), mean average scale error (mASE), mean average orientation error (mAOE), and mean average velocity error (mAVE). Furthermore, to comprehensively evaluate the detection performance, we defined the OmniHD-Scenes detection score (ODS) as follows.

\begin{equation}
ODS = \frac{1}{8}[4mAP + \sum_{mTP\in TP} (1-\min(1,mTP))]
\end{equation}

The ODS score was equally weighted between the detection capability (mAP) and detection accuracy metrics from multiple aspects, including the distance, size, orientation, and velocity. The mAVE, mAOE, and mATE were restricted to the range of 0--1 through application of min(1,mTP).

\subsubsection{Baselines and Implementation Details}

\begin{figure}
    \centering
    \includegraphics[width=\linewidth]{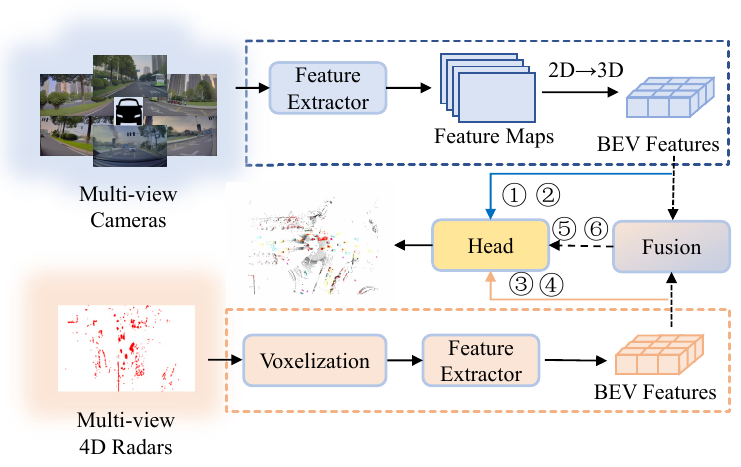}
    \vspace{-0.6cm}
    \caption{Architecture overview of BEV-based 3D object detection baselines. The pipeline processed multi-view camera images and 4D-radar point clouds to generate BEV features, which were then either directly input to detection heads or processed through fusion modules for object attribute prediction. The numbered components represent different baseline methods: {\fontsize{9pt}{9pt}\selectfont \ding{172},\ding{173}} represent LSS \cite{philion2020lift} and BEVFormer r\cite{li2022bevformer}, respectively; {\fontsize{9pt}{9pt}\selectfont \ding{174},\ding{175}} denote PointPillars \cite{lang2019pointpillars} and RadarPillarNet \cite{zheng2023rcfusion}, respectively; whereas {\fontsize{9pt}{9pt}\selectfont \ding{176},\ding{177}} correspond to BEVFusion \cite{liang2022bevfusion} and RCFusion \cite{zheng2023rcfusion}, respectively.}
    \label{fig:bev-objectdet}
\end{figure}

We implemented several established BEV detection methods on OmniHD-Scenes, as illustrated in Fig. \ref{fig:bev-objectdet}. The fundamental principle of BEV-based detection methods is that the raw sensor data are first transformed into BEV representations. The object attributes are then predicted using the detection head.
As camera-only methods, we adopted LSS \cite{philion2020lift} and BEVFormer \cite{li2022bevformer}. For LSS, the settings reported in \cite{liang2022bevfusion} were employed. Additionally, we incorporated DepthNet for depth estimation and used the LiDAR depth for explicit supervision. For BEVFormer, the settings in \cite{li2022bevformer} were employed. Furthermore, we added a temporal version with three frames and experimented with a larger image resolution input (864 $\times$ 1536) and a larger backbone (R101-DCN). The pre-trained weights of both backbones were from FCOS3D \cite{wang2021fcos3d}, consistent with \cite{li2022bevformer}. Moreover, both were trained for 24 epochs using the AdamW optimizer with an initial learning rate of 2e-4. The BEV feature map size was 160 $\times$ 240.

As the 4D-radar baseline, we chose pillar-based methods; namely, PointPillars \cite{lang2019pointpillars} and RadarPillarNet \cite{zheng2023rcfusion}. The input point-cloud range was \{$(x, y, z)$$| $$-60m < x < 60m$, $-40m < y < 40m$, 
$-3m < z < 5m$\}. The input raw radar point-cloud features $f_{in}$ were as follows:
\begin{equation}
	f_{in} = [x, y, z, Power,SNR, v_{xr}, v_{yr}, t_{diff}]
\end{equation}
where $x, y, z$ are the coordinates; $ v_{xr} $ and $ v_{yr} $ are the radial velocity components compensated for ego-motion; $Power$ and $SNR$ are the amplitude and signal-to-noise ratio of the radar point cloud, respectively; and $t_{diff}$ is the time difference from the first frame. By default, we used three frames of accumulated radar points as input. The pillar size for both was $[0.25, 0.25, -8]m$, and both were trained using the AdamW optimizer with an initial learning rate of 1e-3 for 24 epochs.

As fusion methods, we selected BEVFusion \cite{liang2022bevfusion} and RCFusion\cite{zheng2023rcfusion}. We use the LSS pre-trained weights for the image branch, and PointPillars and RadarPillarNet for the point-cloud branch, respectively. They were jointly trained end-to-end for 12 epochs using the AdamW optimizer with an initial learning rate of 2e-4.

Additionally, we implemented a LiDAR baseline. Its model configuration was essentially identical to the radar input. For the 32-beam baseline model, we downsampled from the original 128-beam point cloud based on the LiDAR elevation angle information.

\subsubsection{Results and Analysis}

\input{tables/OD-results}

\input{tables/OD-results-badcondition}

Table \ref{tab:od_performance} compares various sensor modalities and methods for 3D object detection on the OmniHD-Scenes test set. The methods are categorized by input modality into LiDAR-, 4D radar-, camera-, and fusion-based approaches.

For the LiDAR-based methods, we implemented PointPillars \cite{lang2019pointpillars}. With 128 beams, this approach achieved 61.15 mAP and 55.54 ODS, outperforming other input modalities. The advantage of this method arises from the high resolution and point-cloud density provided by high-beam LiDAR, which facilitates higher-precision geometric object representation. Notably, when the LiDAR beams were downsampled to 32, the performance declined (-3.91 mAP, -2.88 ODS). This result highlights the critical role of point-cloud density in detector performance. However, as the LiDAR beams were not uniformly distributed, the downsampling retained most of the object contour information within the annotated range, and maintained an acceptable detection performance.

Among the 4D radar-based methods, RadarPillarNet \cite{zheng2023rcfusion} exhibited slightly better performance than PointPillars \cite{lang2019pointpillars} (+1.06 mAP, +0.60 ODS). This higher performance was primarily due to the hierarchical feature extraction performed by RadarPillarNet, which more effectively captures 4D-radar characteristics. However, a significant performance gap remained compared to the LiDAR result, which was influenced by the sparsity and noise of the 4D-radar point clouds. Notably, a significant improvement in velocity error was observed for the 4D radar results compared to those obtained via LiDAR (0.6982 vs. 1.8731 mAVE, respectively), demonstrating that the radar Doppler information provides precise velocity measurements, which is particularly beneficial for estimating the speed of dynamic objects.

Among the camera-based methods, BEVFormer \cite{li2022bevformer} achieved superior results than LSS \cite{philion2020lift}. With an input resolution of 544 × 960 and a ResNet50 backbone, BEVFormer achieved 26.49 mAP, with further performance gains following the incorporation temporal information (+2.68 mAP, +2.44 ODS). This demonstrates the effectiveness of this method in modeling BEV scenes through spatial and temporal attention mechanisms. Additionally, experiments with higher input resolution (864 × 1536) and a larger backbone (ResNet101-DCN) yielded further performance gains. Notably, although the vision-based methods outperformed the 4D-radar inputs in terms of mAP, they underperformed the 4D-radar methods in terms of mATE and mAVE.

Among the methods based on 4D-radar/camera fusion, RCFusion \cite{zheng2023rcfusion} surpassed BEVFusion \cite{liang2022bevfusion} across various metrics. Its interactive attention mechanism adaptively fused the geometric and semantic features from both modalities; however, a slight decrease in mAVE performance was observed. Compared to the methods using either vision or 4D radar alone, this modal fusion approach generally demonstrated higher overall performance.

Additionally, we evaluated clips from the test set that were specifically labeled as featuring night and rainy conditions, as reported in Table \ref{tab:od_bad_performance}. A comparison with Table \ref{tab:od_performance} reveals a performance drop for the vision-based methods in cases of adverse weather conditions (-1.23 mAP to -1.99 mAP); however, the methods based on 4D radar or 4D-radar/camera fusion showed improvement (+1.48 to +4.90 mAP). Therefore, 4D radar exhibits greater robustness and interference resistance under challenging conditions, which is a crucial feature in the context of reliable autonomous driving systems.

Overall, LiDAR-based methods continued to exhibit superior detection accuracy compared to those of other modal inputs. However, low-cost cameras and radar can narrow this gap through complementary modalities. Moreover, 4D imaging radar demonstrated unique advantages in terms of range and velocity measurements, while exhibiting higher robustness. This performance will inspire researchers to develop improved algorithms that can fully leverage the strengths of 4D radar and achieve more effective fusion with other modalities.

\subsection{3D Semantic Occupancy}
\begin{figure}
    \centering
    \includegraphics[width=\linewidth]{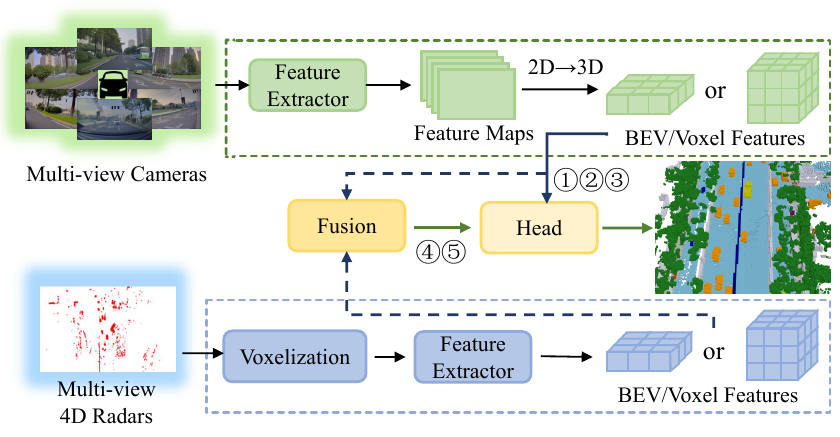}
    \vspace{-0.6cm}
    \caption{Architectural overview of BEV- and voxel-based 3D occupancy prediction baselines. Multi-view images were processed for BEV or voxel feature extraction; these were either fed directly into the occupancy head or first combined with 4D-radar point-cloud features for prediction. The numbered components represent different baseline methods: {\fontsize{9pt}{9pt}\selectfont \ding{172},\ding{173},\ding{174}} represent C-CONet \cite{wang2023openoccupancy}, SurroundOcc \cite{wei2023surroundocc}, and BEVFormer \cite{li2022bevformer}, respectively; and {\fontsize{9pt}{9pt}\selectfont \ding{175},\ding{176}} denote BEVFusion \cite{liang2022bevfusion} and M-CONet \cite{wang2023openoccupancy} respectively.}
    \label{fig:architectural_occ}
\end{figure}

\subsubsection{Evaluation Metrics}
For 3D occupancy prediction, performance metrics combine the occupancy state and semantic segmentation evaluations. Here, we used the intersection over union (IoU) \cite{behley2019semantickitti} as the semantic metric for each category, with the mean IoU (mIoU) averaged across all categories to assess the overall performance. The calculation is expressed as follows:

\begin{equation}
    \mathrm{IoU} =\frac{T P}{T P+F P+F N}
\end{equation}
\begin{equation}
    \mathrm{mIoU} = \frac{1}{C} \sum_{i = 1}^{C} \frac{T P_{i}}{T P_{i}+F P_{i}+F N_{i}}
\end{equation}
where $TP$, $FP$, and $FN$ represent the true-positive, false-positive, and false-negative prediction counts, respectively. Further, $C$ denotes the number of categories, excluding free space. Additionally, we assessed the occupancy-state estimation accuracy using the scene completion IoU (SC IoU), which evaluates geometric accuracy by computing IoU metrics for free space and occupied elements.

\subsubsection{Baselines and Implementation Details}
We implemented classical BEV- and voxel-based occupancy methods on OmniHD-Scenes, as illustrated in Fig. \ref{fig:architectural_occ}. The input specifications, such as the point-cloud range and attributes, were consistent with those used in the object detection task.

For the camera-only approach, we adapted C-CONet \cite{wang2023openoccupancy}, SurroundOcc \cite{wei2023surroundocc}, and BEVFormer \cite{li2022bevformer}. For C-CONet, the settings reported in \cite{wang2023openoccupancy} were employed and the coarse-to-fine ratio was set to 2. For BEVFormer, we evaluated a temporal enhanced version to assess the effectiveness of its temporal modeling capabilities. We also explored performance enhancements by using a higher image resolution (864 $\times$ 1536) and stronger backbone network (R101-DCN). Following \cite{li2022bevformer}, we utilized FCOS3D \cite{wang2021fcos3d} pre-trained models for all three backbones. All models were trained for 24 epochs using the AdamW optimizer with an initial learning rate of 2e-4.

We also implemented fusion-based baseline models that combined camera and 4D-radar data. BEVFusion \cite{liang2022bevfusion} served as a baseline, and was modified by replacing the detection head with an occupancy prediction head. M-CONet \cite{wang2023openoccupancy} was also employed; this is an extension of CONet that incorporates 4D-radar voxel feature extraction with adaptive fusion capabilities. Both models were trained end-to-end for 24 epochs using the AdamW optimizer with an initial learning rate of 2e-4.

\subsubsection{Results and Analysis}

\input{tables/Occ_result}
\input{tables/Occ_result_badcondition}

Tables \ref{tab:occ_performance} and \ref{tab:occ_performance_bad} present the evaluation results for the various methods when implemented on the OmniHD-Scenes test set for occupancy prediction. Both standard and adverse conditions were considered. 

Among the camera-based methods, BEVFormer-T \cite{li2022bevformer} achieved the best results compared to the other baselines. With an input resolution of 544 $\times$ 960 and a ResNet50 backbone, it achieved 16.23 mIoU and 28.42 SC IoU, surpassing the baseline BEVFormer (+1.26 mIoU) and SurroundOcc \cite{wei2023surroundocc} (+1.03 mIoU). C-CONet exhibited poorer performance (13.42 mIoU, 25.69 SC IoU). This demonstrates the effectiveness of temporal information integration in occupancy prediction. Experiments performed with a higher input resolution (864 $\times$ 1536) and larger backbone (ResNet101-DCN) yielded further performance gains, with BEVFormer achieving 16.41 mIoU and 28.30 SC IoU, while BEVFormer-T achieved 17.49 mIoU and 29.74 SC IoU. These results demonstrate that enhanced model capacity benefits fine-grained occupancy prediction.

Among the methods based on 4D-radar/camera fusion, BEVFusion \cite{liang2022bevfusion} and M-CONet \cite{wang2023openoccupancy} exhibited improved mIoU metrics compared to those of the single-frame camera-based methods. As reported in Table \ref{tab:occ_performance}, BEVFusion and M-CONet achieved 16.24 and 16.08 mIoU, respectively. However, when the image resolution and backbone were increased, BEVFormer exhibited superior performance, surpassing BEVFusion (17.49 vs. 16.24 mIoU). This suggests that current fusion approaches have not fully leveraged the unique properties of 4D imaging radar. Therefore, researchers should explore more effective fusion methods to unlock the potential of 4D radar for occupancy prediction.

Furthermore, as reported in Table \ref{tab:occ_performance_bad}, all baseline methods exhibited performance degradation under adverse conditions. Compared to the results listed in Table \ref{tab:occ_performance}, the camera-only approaches exhibited the most significant mIoU decline (1.1-1.65 mIoU). The 4D-radar/camera fusion methods (BEVFusion and M-CONet) demonstrated greater stability with smaller decreases in mIoU (0.78-0.88 mIoU). Moreover, under identical input-resolution and backbone configurations, the fusion methods consistently outperformed the camera-only approaches in terms of mIoU. These results underscore the inherent robustness of 4D radar when applied to adverse conditions.

In summary, the camera-only methods such as BEVFormer-T exhibited strong performance with high-resolution inputs and temporal information integration, achieving the highest mIoU of 17.49 under standard conditions. Further, the fusion-based methods (BEVFusion and M-CONet) exhibited competitive performance and superior robustness under adverse conditions, with smaller performance drops. Notably, their current performance suggests potential for improvement. These findings indicate the need for more advanced fusion strategies to better leverage the complementary strengths of 4D radar and cameras.

%% file: tables/OD-results.tex
\captionsetup{font=small}  

\setlength{\tabcolsep}{1.4pt}  
\renewcommand{\arraystretch}{1}  

\begin{table*}[t]
\centering
\caption{Comparison of 3D object detection on the OmniHD-Scenes \emph{test} set. ``L'' denotes LiDAR, ``C'' denotes camera and ``R'' denotes 4D radar. ``-128'' and ``-32'' represent 128 lines and manually downsampled 32 lines, respectively.}
\resizebox{0.9\textwidth}{!}{
\begin{tabular}{l|c|c|c|cc|cccccc}
\hline\noalign{\smallskip}
Methods &   Image Res.& Modality &   Backbone & mAP$\uparrow$ & ODS$\uparrow$ & mATE$\downarrow$ &mASE$\downarrow$ &mAOE$\downarrow$  &mAVE$\downarrow$ \\
\noalign{\smallskip}
\hline
\noalign{\smallskip}
PointPillars-128~\cite{lang2019pointpillars}      &-          & L &-   &61.15    & 55.54 &   0.2825   &    0.1981   &     0.5222    &     1.8763    \\
PointPillars-32~\cite{lang2019pointpillars}        &-         & L   &-  &57.24   & 52.66 &  0.3040    &    0.2038   &     0.5692    &  1.8731       \\

\noalign{\smallskip}
\hline
\noalign{\smallskip}
\noalign{\smallskip}
\hline
\noalign{\smallskip}
PointPillars~\cite{lang2019pointpillars}           &-       & R  &-  &23.82    & 37.21 &   0.6752   &    0.2447   &     0.3776    &     0.6789    \\
RadarPillarNet~\cite{zheng2023rcfusion}           &-      & R &-    &24.88   & 37.81 &  0.6597    &    0.2389   &     0.3736    &  0.6982       \\ 
\noalign{\smallskip}
\hline
\noalign{\smallskip}

LSS~\cite{philion2020lift}           &544×960       & C   &R50     & 22.44 & 26.01 & 1.0238 & 0.2230   & 0.5942 & 2.0138   \\

BEVformer~\cite{li2022bevformer}       &544×960        & C     &R50   & 26.49 & 28.10 & 1.1430 & 0.2315   & 0.5799 & 1.6666  \\
BEVformer-T~\cite{li2022bevformer}           &544×960      & C      &R50  & 29.17 & 30.54 & 1.1046 & 0.2346   & 0.4889 & 1.0797   \\
BEVformer~\cite{li2022bevformer}       &864×1536        & C     &R101-DCN   & 30.10 & 30.55 & 1.0633 & 0.2266   & 0.5331 & 1.6625  \\
BEVformer-T~\cite{li2022bevformer}           &864×1536       & C      &R101-DCN  & 32.22 & 32.57 & 1.0637 & 0.2271   & 0.4558 & 1.0683   \\

\noalign{\smallskip}
\hline
\noalign{\smallskip}
BEVFusion~\cite{liang2022bevfusion}          &544×960     & C\&R    &R50    & 33.95 & 43.00 & 0.5703 & 0.2165   & 0.3814 & 0.7474  \\
RCFusion~\cite{zheng2023rcfusion}              &544×960   & C\&R    &R50    & 34.88 & 41.53 & 0.5676 & 0.2135   & 0.3711 & 0.9208   \\

\hline
\end{tabular}}
\label{tab:od_performance}
\end{table*}
\setlength{\tabcolsep}{1.4pt}

%% file: tables/OD-results-badcondition.tex
\captionsetup{font=small}  

\setlength{\tabcolsep}{1.4pt}  
\renewcommand{\arraystretch}{1}  

\begin{table*}[t]
\centering
\caption{Comparison of 3D object detection under adverse conditions (night and rainy weather) on the OmniHD-Scenes \emph{test} set.}
\begin{adjustbox}{width=0.9\textwidth,center}
\begin{tabular}{l|c|c|c|cc|cccccc}
\hline\noalign{\smallskip}
Methods & Image Res.&  Modality &   Backbone & mAP$\uparrow$ & ODS$\uparrow$ & mATE$\downarrow$ &mASE$\downarrow$ &mAOE$\downarrow$  &mAVE$\downarrow$ \\
\noalign{\smallskip}
\hline
\noalign{\smallskip}
PointPillars-128~\cite{lang2019pointpillars}             &-   & L &-   &60.02    & 55.03 &   0.2889   &    0.2008   &     0.5084    &    1.8996    \\
PointPillars-32~\cite{lang2019pointpillars}          &-      & L   &-  &57.09   & 52.94 &  0.3014    &    0.2098   &     0.5370    &  1.9397       \\

\noalign{\smallskip}
\hline
\noalign{\smallskip}
\noalign{\smallskip}
\hline
\noalign{\smallskip}
PointPillars~\cite{lang2019pointpillars}          &-       & R  &-  &28.63    & 40.70 &   0.6539   &    0.2538   &     0.3703    &     0.6112    \\
RadarPillarNet~\cite{zheng2023rcfusion}         &-       & R &-    &29.78   & 41.00 &  0.6473    &   0.2482   &   0.3707   &  0.6447      \\ 
\noalign{\smallskip}
\hline
\noalign{\smallskip}

LSS~\cite{philion2020lift}         &544×960         & C   &R50     & 21.21 & 25.62 & 1.0723 & 0.2373   & 0.5615 & 1.8993   \\

BEVformer~\cite{li2022bevformer}           &544×960    & C     &R50   & 24.57 & 27.25 & 1.1603 & 0.2449   & 0.5581 & 1.6911  \\
BEVformer-T~\cite{li2022bevformer}           &544×960      & C      &R50  & 27.50 & 29.92 & 1.1038 & 0.2477   & 0.4588 & 1.0685   \\
BEVformer~\cite{li2022bevformer}            &864×1536     & C      &R101-DCN  & 28.11 & 29.36 & 1.0972 & 0.2400   & 0.5358 & 1.6947   \\
BEVformer-T~\cite{li2022bevformer}           &864×1536      & C      &R101-DCN  &30.39 & 31.61 & 1.0613 & 0.2392   & 0.4475 & 1.0780   \\

\noalign{\smallskip}
\hline
\noalign{\smallskip}
BEVFusion~\cite{liang2022bevfusion}          &544×960     & C\&R    &R50    & 35.83 & 44.95 & 0.5665 & 0.2317   & 0.3739 & 0.6653  \\
RCFusion~\cite{zheng2023rcfusion}             &544×960    & C\&R    &R50    & 36.36 & 43.29 & 0.5508 & 0.2283   & 0.3616 & 0.8504   \\

\hline
\end{tabular}
\label{tab:od_bad_performance}
\end{adjustbox}
\end{table*}
\setlength{\tabcolsep}{1.4pt}

%% file: tables/Occ_Result.tex
\definecolor{ncar}{RGB}{255, 165, 0}
\definecolor{npedestrian}{RGB}{128, 0, 128}
\definecolor{nrider}{RGB}{0, 0, 200}
\definecolor{nlarge_vehicle}{RGB}{220, 220, 0}
\definecolor{ncycle}{RGB}{230 ,230 ,230}
\definecolor{nroad_obstacle}{RGB}{255, 69, 0}
\definecolor{ntraffic_fence}{RGB}{0, 0, 150}
\definecolor{ndriveable_surface}{RGB}{135 ,206 ,235}
\definecolor{nsidewalk}{RGB}{200, 200, 200}
\definecolor{nvegetation}{RGB}{34 ,139 ,34}
\definecolor{nmanmade}{RGB}{230, 230, 250}

\captionsetup{font=small}  

\setlength{\tabcolsep}{1.4pt}  
\renewcommand{\arraystretch}{1}  

\begin{table*}[t]
\centering
\caption{Comparison of occupancy prediction on the OmniHD-Scenes \emph{test} set. ``C'' denotes camera and ``R'' denotes 4D radar. 
}
\begin{adjustbox}{width=0.9\textwidth,center}
\begin{tabular}{l|c|c|c|cc|ccccccccccc}
\hline
\noalign{\smallskip}
Methods     & Image Res. & Modality & Backbone & SC IoU & mIoU   & \rotatebox{90}{\textcolor{ncar}{$\blacksquare$}car}    & \rotatebox{90}{\textcolor{npedestrian}{$\blacksquare$}pedestrian} & \rotatebox{90}{\textcolor{nrider}{$\blacksquare$}rider} & \rotatebox{90}{\textcolor{nlarge_vehicle}{$\blacksquare$}large vehicle} & \rotatebox{90}{\textcolor{ncycle}{$\blacksquare$}cycle} & \rotatebox{90}{\textcolor{nroad_obstacle}{$\blacksquare$}road obstacle} & \rotatebox{90}{\textcolor{ntraffic_fence}{$\blacksquare$}traffic fence} & \rotatebox{90}{\textcolor{ndriveable_surface}{$\blacksquare$}drive. surf.} & \rotatebox{90}{\textcolor{nsidewalk}{$\blacksquare$}sidewalk} & \rotatebox{90}{\textcolor{nvegetation}{$\blacksquare$}vegetation} & \rotatebox{90}{\textcolor{nmanmade}{$\blacksquare$}manmade} \\
\noalign{\smallskip}
\hline
\noalign{\smallskip}
C-CONet~\cite{wang2023openoccupancy}     & 544×960  & C    & R50  & 25.69 & 13.42 & 20.03 & 3.51 & 11.71 & 16.62 & 0.79 & 1.14 & 22.75 & 33.57 & 14.82 & 17.73 & 4.93 \\
SurroundOcc~\cite{wei2023surroundocc} & 544×960  & C    & R50  & 28.61 & 15.20 & 21.46 & 3.96 & 10.76 & 16.58 & 1.57 & 2.99 & 21.63 & 48.52 & 18.31 & 16.73 & 4.71 \\
BEVFormer~\cite{li2022bevformer}   & 544×960  & C    & R50  & 27.04 & 14.97 & 20.64 & 5.87 & 14.40 & 16.68 & 1.52 & 3.64 & 20.64 & 46.61 & 16.19 & 14.80 & 3.69 \\
BEVFormer-T~\cite{li2022bevformer} & 544×960  & C    & R50  & 28.42 & 16.23 & 22.73 & 5.45 & 14.7  & 18.21 & 3.09 & 3.87 & 21.54 & 48.15 & 17.58 & 17.77 & 5.48 \\
\noalign{\smallskip}
\hline
\noalign{\smallskip}
BEVFormer~\cite{li2022bevformer}   & 864×1536 & C    & R101-DCN & 28.30 & 16.41 & 23.72 & 6.37 & 16.33 & 20.44 & 1.78 & 3.78 & 22.21 & 48.55 & 17.88 & 15.49 & 3.99 \\
BEVFormer-T~\cite{li2022bevformer} & 864×1536 & C    & R101-DCN & 29.74 & 17.49 & 24.90 & 6.48 & 16.45 & 21.49 & 2.87 & 4.62 & 22.51 & 49.92 & 18.59 & 18.53 & 5.96 \\
\noalign{\smallskip}
\hline
\noalign{\smallskip}
BEVFusion~\cite{liang2022bevfusion}   & 544×960  & C\&R & R50  & 27.02 & 16.24 & 27.02 & 4.78 & 21.71 & 21.59 & 1.55 & 2.78 & 25.21 & 44.35 & 12.32 & 13.06 & 4.25 \\
M-CONet~\cite{wang2023openoccupancy}     & 544×960  & C\&R & R50  & 27.74 & 16.08 & 25.21 & 3.42 & 17.53 & 21.46 & 0.88 & 0.58 & 29.88 & 34.48 & 14.89 & 19.57 & 8.98 \\
\noalign{\smallskip}
\hline
\end{tabular}
\label{tab:occ_performance}
\end{adjustbox}
\end{table*}
\setlength{\tabcolsep}{1.4pt}

%% file: tables/Occ_result_badcondition.tex
\definecolor{ncar}{RGB}{255, 165, 0}
\definecolor{npedestrian}{RGB}{128, 0, 128}
\definecolor{nrider}{RGB}{0, 0, 200}
\definecolor{nlarge_vehicle}{RGB}{220, 220, 0}
\definecolor{ncycle}{RGB}{230 ,230 ,230}
\definecolor{nroad_obstacle}{RGB}{255, 69, 0}
\definecolor{ntraffic_fence}{RGB}{0, 0, 150}
\definecolor{ndriveable_surface}{RGB}{135 ,206 ,235}
\definecolor{nsidewalk}{RGB}{200, 200, 200}
\definecolor{nvegetation}{RGB}{34 ,139 ,34}
\definecolor{nmanmade}{RGB}{230, 230, 250}

\captionsetup{font=small}  

\setlength{\tabcolsep}{1.4pt}  
\renewcommand{\arraystretch}{1}  

\begin{table*}[t]
\centering
\caption{Comparison of occupancy prediction under adverse conditions (night and rainy weather) on the OmniHD-Scenes \emph{test} set.  
}
\begin{adjustbox}{width=0.9\textwidth,center}
\begin{tabular}{l|c|c|c|cc|ccccccccccc}
\hline
\noalign{\smallskip}
Methods     & Image Res. & Modality & Backbone & SC IoU & mIoU   & \rotatebox{90}{\textcolor{ncar}{$\blacksquare$}car}    & \rotatebox{90}{\textcolor{npedestrian}{$\blacksquare$}pedestrian} & \rotatebox{90}{\textcolor{nrider}{$\blacksquare$}rider} & \rotatebox{90}{\textcolor{nlarge_vehicle}{$\blacksquare$}large vehicle} & \rotatebox{90}{\textcolor{ncycle}{$\blacksquare$}cycle} & \rotatebox{90}{\textcolor{nroad_obstacle}{$\blacksquare$}road obstacle} & \rotatebox{90}{\textcolor{ntraffic_fence}{$\blacksquare$}traffic fence} & \rotatebox{90}{\textcolor{ndriveable_surface}{$\blacksquare$}drive. surf.} & \rotatebox{90}{\textcolor{nsidewalk}{$\blacksquare$}sidewalk} & \rotatebox{90}{\textcolor{nvegetation}{$\blacksquare$}vegetation} & \rotatebox{90}{\textcolor{nmanmade}{$\blacksquare$}manmade} \\
\noalign{\smallskip}
\hline
\noalign{\smallskip}
C-CONet~\cite{wang2023openoccupancy}     & 544×960  & C    & R50  & 24.68 & 12.32 & 19.62 & 3.03 & 10.03 & 15.00 & 1.08 & 0.51 & 16.18 & 31.83 & 15.23 & 17.48 & 5.47 \\
SurroundOcc~\cite{wei2023surroundocc} & 544×960  & C    & R50  & 27.20 & 14.04 & 21.07 & 3.78 & 9.56  & 15.03 & 1.97 & 2.41 & 14.85 & 45.58 & 18.63 & 16.07 & 5.43 \\
BEVFormer~\cite{li2022bevformer}   & 544×960  & C    & R50  & 25.55 & 13.65 & 20.60 & 5.41 & 11.99 & 15.35 & 2.17 & 2.48 & 13.43 & 43.16 & 16.94 & 14.41 & 4.25 \\
BEVFormer-T~\cite{li2022bevformer} & 544×960  & C    & R50  & 27.08 & 14.93 & 22.68 & 4.66 & 12.27 & 17.23 & 4.11 & 2.59 & 14.87 & 44.88 & 17.23 & 17.33 & 6.41 \\
\noalign{\smallskip}
\hline
\noalign{\smallskip}
BEVFormer~\cite{li2022bevformer}   & 864×1536 & C    & R101-DCN & 26.66 & 15.03 & 23.52 & 5.85 & 13.61 & 19.62 & 2.62 & 2.28 & 14.59 & 45.21 & 18.55 & 14.82 & 4.70 \\
BEVFormer-T~\cite{li2022bevformer} & 864×1536 & C    & R101-DCN & 28.41 & 15.84 & 24.59 & 5.69 & 13.37 & 20.09 & 3.59 & 3.04 & 13.80 & 46.46 & 18.74 & 18.22 & 6.62 \\
\noalign{\smallskip}
\hline
\noalign{\smallskip}
BEVFusion~\cite{liang2022bevfusion}   & 544×960  & C\&R & R50  & 25.32 & 15.36 & 27.17 & 3.58 & 19.17 & 22.62 & 2.31 & 1.55 & 21.72 & 40.93 & 12.54 & 12.64 & 4.69 \\
M-CONet~\cite{wang2023openoccupancy}     & 544×960  & C\&R & R50  & 26.73 & 15.30 & 25.37 & 3.07 & 16.13 & 20.97 & 1.41 & 0.25 & 23.95 & 32.65 & 15.47 & 19.75 & 9.21 \\
\noalign{\smallskip}
\hline
\end{tabular}
\label{tab:occ_performance_bad}
\end{adjustbox}
\end{table*}
\setlength{\tabcolsep}{1.4pt}

%% file: sections/07_conclusion.tex
\section{Conclusion and Future Work}
\label{sec:conclusion}

In this study, we introduced  OmniHD-Scenes, a large-scale, multimodal dataset featuring multi-view cameras and 4D radar to explore cost-effective sensor solutions for autonomous driving. The dataset contains  1,501 clips of complex driving scenarios, including hazardous test cases from closed sites. Using an advanced 4D annotation pipeline, we annotated 200 of these clips with over 514,000 3D tracking boxes and static-scene point-cloud segmentation. Additionally, we introduced a novel 3D semantic occupancy generation pipeline that automatically creates dense occupancy labels. We established benchmarks for 3D object detection and occupancy prediction, with experiments demonstrating the potential of 4D radar in adverse conditions and aiming to inspire more efficient algorithm development.
In the future, we plan to expand the OmniHD-Scenes dataset to support the development and evaluation of next-generation autonomous driving algorithms. Our efforts will be concentrated on the following areas.

\textbf{End-to-End AEB Benchmark.} We plan to establish an end-to-end AEB benchmark focused on safety-critical scenarios. The OmniHD-Scenes dataset includes hazardous test cases from closed test sites, featuring Autonomous Emergency Braking (AEB) scenarios triggered by specific Time-to-Collision (TTC) conditions, which makes an end-to-end AEB benchmark possible. An end-to-end AEB model can make decisions (e.g., braking) directly from raw sensor input, and its performance can be evaluated in controlled, repeatable, safety-critical contexts. Furthermore, this allows for a re-evaluation of the role of 4D radar in the system. The direct and precise velocity measurements it provides are critical for an E2E AEB model to make split-second, accurate decisions, and it can also ensure reliability at night when vision is compromised.

\textbf{Vision-Language-Action (VLA) Benchmark.} We will establish a full-chain, end-to-end benchmark for Vision-Language-Action (VLA) models. We will develop rich instruction sets, including scene description, object queries, occupancy status descriptions, and meta-actions, to ground language commands in the physical world and ultimately achieve end-to-end trajectory planning. Furthermore, this benchmark will encourage researchers to explore deeper multi-modal fusion strategies, particularly how to combine the precise motion information provided by the novel 4D radar to better understand action-related commands, thereby creating a complete evaluation loop from high-level semantic understanding to low-level trajectory planning.